\documentclass[]{fairmeta}

\usepackage{amsmath}
\usepackage{amssymb}
\usepackage{mathtools}
\usepackage{amsthm}
\usepackage{nicefrac}
\usepackage{gensymb}
\usepackage{pifont}
\usepackage{xspace}
\usepackage{adjustbox}
\usepackage{array}
\usepackage{tabularx}
\usepackage{makecell}
\usepackage{enumitem}
\usepackage{rotating}
\usepackage{wrapfig}
\usepackage{appendix}
\usepackage{colortbl}
\def\eg{\emph{e.g.}}

\def\ie{\emph{i.e.}}

\def\etc{\emph{etc.}}

\usepackage{amsmath,amsfonts,bm}

\def\eqref#1{equation~\ref{#1}}
\def\1{\bm{1}}

\DeclareMathAlphabet{\mathsfit}{\encodingdefault}{\sfdefault}{m}{sl}
\SetMathAlphabet{\mathsfit}{bold}{\encodingdefault}{\sfdefault}{bx}{n}

\newcommand{\lr}{\alpha}

\title{VCN-Bench: A Video-Contextualized Navigation Benchmark for Spatial Reasoning over Prior Visual Experience}

\author[1,*]{Siqi Zhang}
\author[3]{Meng Wei}
\author[4]{Chenyang Wan}
\author[4]{Shaohao Zhu}
\author[5]{Shufan Shen}
\author[3]{Xihui Liu}
\author[1,\dagger]{Zhihua Wei}
\author[2,\dagger]{Tai Wang}
\author[2]{Jiangmiao Pang}

\affiliation[1]{Tongji University}
\affiliation[2]{Shanghai AI Laboratory}
\affiliation[3]{The University of Hong Kong}
\affiliation[4]{Zhejiang University}
\affiliation[5]{Institute of Computing Technology, Chinese Academy of Sciences}

\contribution[*]{Work was done during internship at Shanghai AI Laboratory.}
\contribution[\dagger]{Corresponding Authors.}

\hypersetup{
  pdftitle={VCN-Bench: A Video-Contextualized Navigation Benchmark for Spatial Reasoning over Prior Visual Experience},
  pdfauthor={Siqi Zhang, Meng Wei, Chenyang Wan, Shaohao Zhu, Shufan Shen, Xihui Liu, Zhihua Wei, Tai Wang, Jiangmiao Pang},
  pdfsubject={Video-contextualized navigation benchmark for closed-loop spatial reasoning},
  pdfkeywords={embodied navigation, spatial reasoning, multimodal large language models, video-contextualized navigation, Matterport3D}
}

\metadata[Code]{\url{https://github.com/siqiZ805/VCN-Bench.git}}
\metadata[Benchmark]{\url{https://huggingface.co/datasets/qqqqi/VSI-NavBench}}

\abstract{
Spatial reasoning is fundamental to embodied agents, yet it remains unclear whether spatial understanding can be carried forward to guide sequential interactions.
Existing spatial-reasoning benchmarks typically terminate at offline predictions, while navigation benchmarks evaluate spatial reasoning as part of instruction following and exploration.
We introduce VCN-Bench, a \textbf{V}ideo-\textbf{C}ontextualized \textbf{N}avigation benchmark for probing closed-loop spatial reasoning over prior visual experience in MLLMs. Given a prior video covering both the initial location and destination, the agent is tasked with reasoning out the instruction-specified target and navigating toward it with the inferred spatial context.
Built on Matterport3D, VCN-Bench contains five instruction types, 100k training episodes, and 1,250 evaluation episodes. Navigation serves as the primary evaluation, while diagnostic goal identification helps distinguish destination-resolution errors from subsequent navigation failures.
We further propose MV-DualVLN, a planning-oriented baseline that jointly leverages prior video and in-episode observations. Experiments reveal limited navigation performance, a substantial destination-resolution-to-navigation gap, and frequent navigation failures even after correct destination identification.
}

\begin{document}

\maketitle

\begin{figure}[t]
    \centering
    \includegraphics[width=\linewidth]{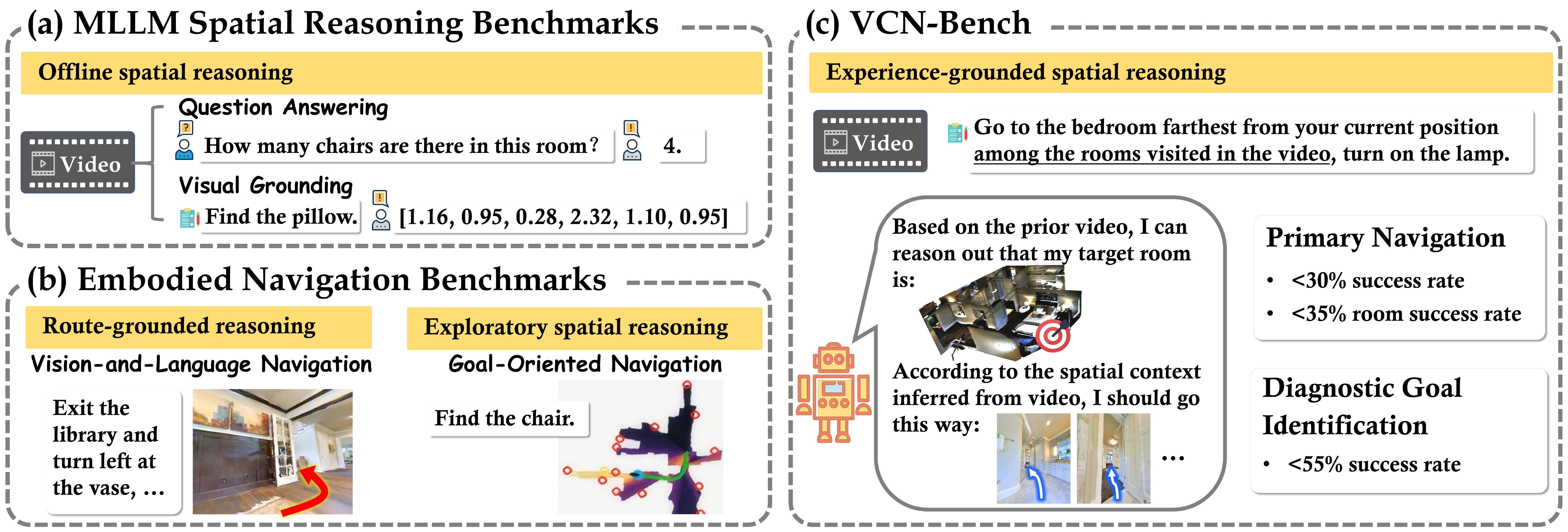}
    \caption{VCN-Bench is a video-contextualized navigation benchmark tailored for probing closed-loop spatial reasoning over prior visual experience in MLLMs. The spatial context inferred from the prior video directly informs destination resolution and guides closed-loop action decisions.}
    \label{fig:teaser}
\end{figure}

\section{Introduction}

Spatial reasoning ability is fundamental to embodied agents, enabling them to understand the relationships among locations, regions, and objects and to use this understanding to guide interaction in physical environments~\citep{zhu2026vst}. 
As multimodal large language models (MLLMs) increasingly serve as reasoning backbones for embodied agents~\citep{zhang2024uninavid,zheng2025multimodal,wang2026qwenvla}, it is important to evaluate their spatial reasoning capabilities in embodied settings, \ie, whether MLLMs can infer spatial relationships from visual observations and leverage such spatial understanding to support subsequent interactions.
% As multimodal large language models (MLLMs) increasingly serve as reasoning backbones for embodied agents~\citep{zhang2024uninavid,zheng2025multimodal}, it is important to evaluate not only whether MLLMs can infer spatial relationships from visual observations, but also whether such spatial understanding can support subsequent interaction.

Prior benchmarks evaluate spatial reasoning in MLLMs through question answering and grounding over multi-view observations~\citep{zhang2025spar,yang2025mmsi} and videos~\citep{yang2025vsi,lin2025mmsi-video}. These benchmarks provide measurements of spatial understanding in MLLMs, but typically terminate at an offline prediction. They therefore do not determine whether the inferred spatial context can guide interaction, where actions continually change the agent's observations and the resulting visual feedback informs later decisions.

Embodied navigation complements these benchmarks by evaluating agents through closed-loop interaction.
Vision-and-Language navigation (VLN)~\citep{anderson2018vision,ku2020rxr} provides route-grounded instructions, requiring agents to ground actions and landmarks in the environment.
Goal-oriented navigation~\citep{zhu2017target,chaplot2020object} emphasizes online exploration toward a specified goal, where spatial evidence is progressively acquired through model-dependent exploration.
In these settings, spatial reasoning is executed jointly with instruction following or exploration. This confounding effect makes it difficult to characterize how spatial knowledge is inferred, and how such knowledge is carried forward to guide subsequent navigation.
% observation history is model-dependent with various visual content and scene coverage, coupling spatial-evidence acquisition with how effectively it is reasoned over.
% Despite the substantial progress these paradigms have driven, neither isolates the evaluation of spatial reasoning 
% Therefore, it remains an open question whether an agent can leverage spatial understanding from previous observations to resolve destination and support subsequent navigation.

% Embodied navigation complements these benchmarks by evaluating agents through closed-loop interaction. As illustrated in Figure 1a, vision-and-Language navigation (VLN) evaluates the execution of following landmarks and local actions, while goal-oriented navigation evaluates target search through online exploration. In these settings, navigation failures arise not only from limitations in spatial reasoning over previous observations, but also from landmark following errors and uninformed exploration. Therefore, it remains an open question whether an agent can leverage spatial understanding from previous observations to support subsequent closed-loop navigation.

% Consequently, existing spatial-reasoning benchmarks provide visual context without subsequent action, while navigation benchmarks provide closed-loop action without controlled visual experience. This leaves an important gap between reasoning over a previously observed environment and acting effectively within it.

%%%%%%%%%%%%%%%%%%%%%%%%%%%%%%%%%%%%%%

% vcn-bench：是什么，长什么样，为什么是closed-loop spatial reasoning
To fill this gap, we introduce VCN-Bench, a benchmark for probing closed-loop spatial reasoning over prior visual experience in MLLMs through video-contextualized navigation. 
% 
% We operationalize closed-loop spatial reasoning as the ability to leverage a prior scene video as spatial context to navigate toward the instruction-specified destination in the video through an iterative perception-reasoning-action loop.
% 
As illustrated in \Cref{fig:teaser}, each episode consists of a prior video of an indoor environment and an instruction referring to a room or object observed in the video. 
% The agent is tasked to reason over the prior video to resolve the intended destination and then combine the inferred spatial context with incremental in-episode observations to make sequential navigation decisions.
The prior video provides scene-specific spatial context covering both the initial position and target destination, while the video-grounded instruction makes the inferred spatial knowledge directly shape destination resolution and closed-loop decisions.
% The prior video covers both the agent’s initial location and the target destination while remaining available as scene-specific context throughout the episode. 
% The agent is tasked to reason over it for destination resolution and combine it with incremental egocentric observations to make sequential navigation decisions.
% The video-grounded instruction makes spatial reasoning directly relevant to destination resolution.
In this way, VCN-Bench evaluates not only whether an agent can infer spatial relationships, but also whether the resulting spatial knowledge can be carried forward to support subsequent navigation.
% The video covers both the agent’s initial location and the target destination. 
% This design avoids the confounding effect of existing navigation benchmarks and focuses evaluation on whether information acquired from prior visual experience can support subsequent closed-loop interaction.
% We operationalize closed-loop spatial reasoning as the ability to resolve an instruction-specified destination from prior visual context, relating ongoing observations to that context, and making sequential goal-directed navigation decisions.

%%%%%%%%%%%%%%%%%%%%%%%%%%%%%%%%%%%%%%
% 具体而言，任务，评测
To cover complementary dimensions of spatial reasoning, 
VCN-Bench comprises five instruction types grouped into two goal families. 
Object Goal contains Instance instructions that refer to object instances with unique descriptions.
Room-to-Object Goal requires identifying a target room through Count, Order, Distance, or Orientation relations before specifying an object within it.
% Object Goal instructions refer to an object instance or category observed in the prior video, while Room-to-Object instructions specify a target room before referring to an object within it. These instructions place different demands on spatial memorization and spatial relation understanding.
Video-contextualized navigation is the primary evaluation task. To distinguish whether failures arise from destination resolution or navigation, we additionally introduce a diagnostic goal-identification protocol that asks the model to identify a target room or object in the prior video without navigation. %helping separate destination-identification errors from subsequent planning failures.

%%%%%%%%%%%%%%%%%%%%%%%%%%%%%%%%%%%%%%
% 数据，模型，insight
Built on Matterport3D~\citep{chang2017matterport3d}, VCN-Bench contains 100k training episodes and 1,250 evaluation episodes.
We further introduce MV-DualVLN, an MLLM-based baseline that jointly reasons over the prior video and its own observations to plan actions in pixel space.
% jointly processes the prior video, navigation history, and current multi-view observations for navigation planning.
% predict the next waypoint in pixel-space. Its oracle pixel-goal executor factors out low-level control failures and focuses evaluation on high-level navigation decisions. 
Experiments show that models achieve limited success on video-contextualized navigation. Diagnostic analyses reveal difficulties in destination resolution and show that many correctly identified episodes still fail during navigation, exposing additional challenges in closed-loop reasoning.
% navigation.
% Diagnostic results indicate that models often fail to resolve the correct destination, while a substantial fraction of navigation failures remains after successful destination identification, exposing additional challenges in closed-loop planning. %Ablations further show that the full prior video provides useful spatial context beyond target appearance alone.

%%%%%%%%%%%%%%%%%%%%%%%%%%%%%%%%%%%%%%
% contribution
In summary, our contributions are threefold:
\begin{itemize}[leftmargin=1.0em]
    \vspace{-2pt}
    \item We introduce VCN-Bench, a video-contextualized navigation benchmark with five instruction types for probing closed-loop spatial reasoning over prior visual experience in MLLMs. %It evaluates whether an agent can use a prior scene video and incremental observations to reach an instruction-specified destination.
    \vspace{-2pt}
    \item %We design five instruction types that emphasize entity grounding, trajectory-structured reasoning, and geometry-aware relational reasoning over prior visual experience. 
    We propose a navigation-centered evaluation protocol in which navigation serves as the primary task and goal identification provides a diagnostic probe for distinguishing destination-resolution errors from subsequent navigation failures.
    % We develop a navigation-centered evaluation protocol with six instruction types, each emphasizing different demands on spatial reasoning capabilities. And propose a diagnostic goal-identification probe to help distinguish destination-resolution failures from closed-loop navigation failures.
    \vspace{-2pt}
    \item We introduce MV-DualVLN as a reference baseline for VCN-Bench. Analyses reveal bottlenecks in both destination resolution and navigation, as well as a substantial gap between them.
    % We benchmark representative proprietary and open-source MLLMs and introduce MV-DualVLN as a strong baseline. Our experiments reveal substantial challenges in video-contextualized navigation, and demonstrate the benefit of full prior-video context over target appearance alone.
\end{itemize}
\section{Related Work}

\textbf{Spatial Reasoning Evaluation for MLLMs.}
A growing body of benchmarks has been proposed to evaluate MLLMs' spatial intelligence~\citep{yang2025vsi,yang2025supersensing,hong2026esibench}, including 3D VQA and multi-image spatial reasoning benchmarks~\citep{chen2020scanrefer,azuma2022scanqa,yeshwanth2023scannet++,lyu2024mmscan,ma20253dsrbench,zhang2025spar,yang2025mmsi,lin2025mmsi-video,xu2025multi,li2025sti,huang2026spatialdise}. However, these benchmarks are limited to offline reasoning over fixed inputs in VQA or grounding formats, cannot meet the demand of closed-loop reasoning in physical interaction.
In contrast, VCN-Bench aims to probe closed-loop spatial reasoning in MLLMs through specially designed video-contextualized navigation, where the agent is tasked to navigate within the spatial context implied in the prior video.

\textbf{Navigation Benchmarks.}
As a fundamental task for embodied intelligence, embodied navigation has spurred diverse benchmarks~\citep{zhu2017target,krantz2023navigating,deitke2020robothor,chaplot2020object,savva2019habitat,zhang2025nava3,gao2025octonav}. 
These tasks can be divided into two categories: VLN tasks~\citep{anderson2018vision,ku2020rxr,krantz2020beyond,he2021landmark-rxr,jain2019r4r} focus on instruction fidelity and landmark recognition, while goal-oriented navigation tasks~\citep{qi2020reverie,zhu2021soon,yokoyama2024ovon,huang2025vlln,chaplot2020object,zhu2017target,kadian2020sim2real,zhao2021surprising,partsey2022ismapping} focus on fast exploration via commonsense knowledge.
Recent benchmarks~\citep{wani2020multion,khanna2024goat-bench,song2025lh-vln,hong2025gsa-vln,gao2025octonav} formalize lifelong navigation as a sequence of concatenated episodes, enabling agents to retain and reuse information across episodes.
% enabling cross-episode memory retention.
However, their evaluation of spatial reasoning is confounded with model-dependent factors, such as model-specific exploration and trajectory coverage.
% their prior experience is acquired through model-dependent interaction and therefore varies with exploration and trajectory coverage, confounding subsequent reasoning.
% However, these benchmarks lack controllability over the agent's memory, making it difficult to assess reasoning capabilities over prior experience.
VCN-Bench provides the scene information necessary for navigation tasks in the form of pre-recorded videos, centering evaluation on the agent's ability to reason over prior visual experience and leverage spatial context to guide navigation.

\begin{figure}
    \centering
    \includegraphics[width=0.97\linewidth]{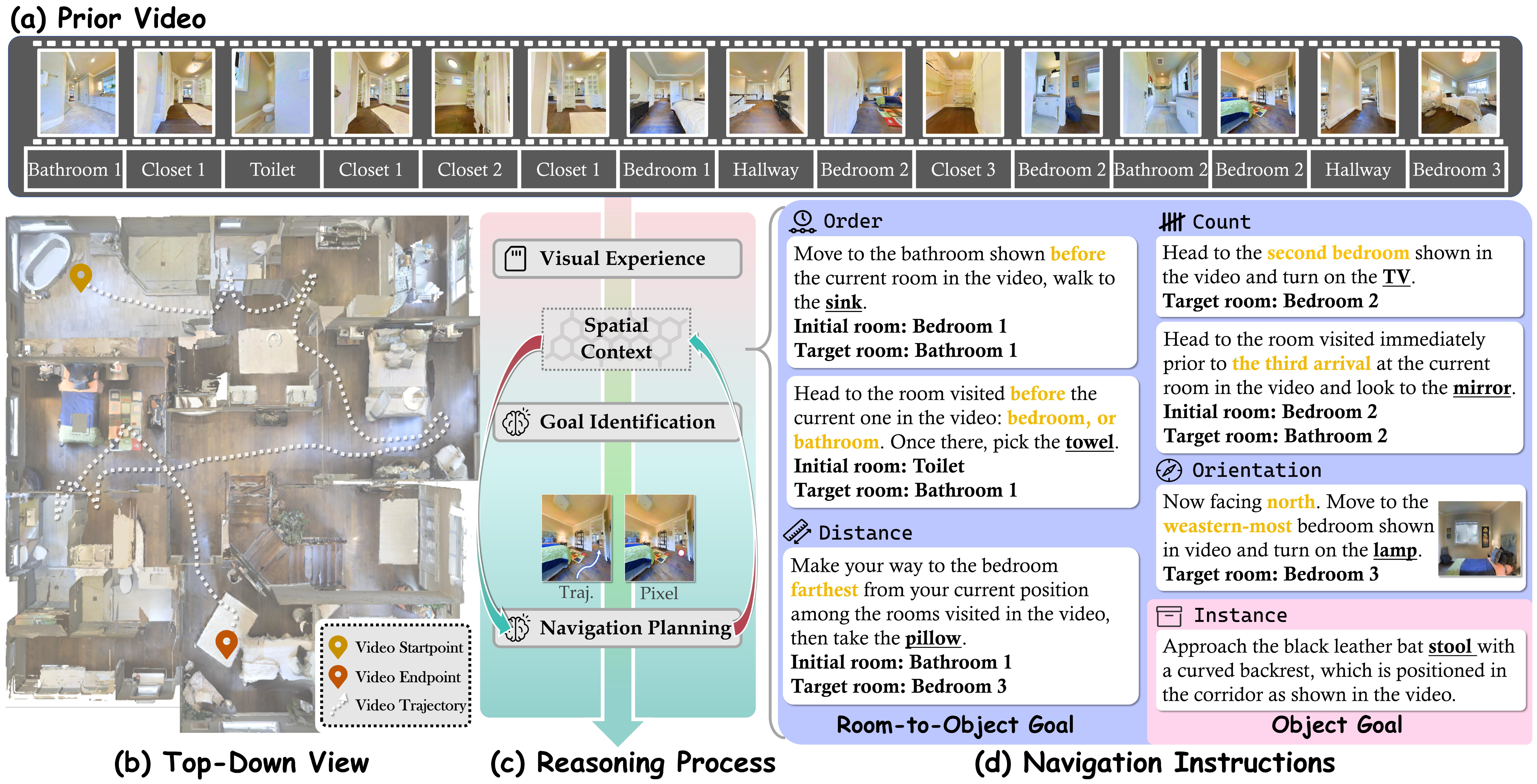}
    \caption{Demonstration of VCN-Bench. (a) Prior video with room labels. (b) Top-down view of the prior video. (c) The reasoning process during navigation. Navigation tasks (d) are conditioned on the prior video with target room labels.}
\end{figure}

\section{VCN-Bench}
\label{sec:vcn-bench}

In this section, we introduce the task formulation (\Cref{sec:task_form}), the design of five distinct instruction types (\Cref{sec:task_design}), and the construction pipeline from videos to episodes (\Cref{sec:bench_construct}) of our VCN-Bench.
Benchmark details are provided in \Cref{app:benchmark}.

\subsection{Task Formulation}
\label{sec:task_form}
In each episode, the agent is initialized in an environment, provided with an RGB-only prior video $\mathcal{V}$, and tasked with a video-contextualized instruction $\mathcal{I}$. 
Evaluation scenes are disjoint from the training scenes. For each evaluation episode, the prior video provides the only scene-specific information available before navigation.
% The agent has no exposure to the environment during training; its scene knowledge comes entirely from the prior video.
The prior video covers both the agent's initial position and the target destination, so there is no need for uninformed exploration of regions outside the video coverage.

% ? action space, stopping criterion
At each step during navigation, the agent can have access to current RGB and depth observations, and current position and heading. Then the agent needs to make a sequence of actions to reach the goal location, where each action $a_t$ belongs to $\mathcal{A}$ = \{FORWARD($0.25m$), TurnLeft($30^{\circ}$), TurnRight($30^{\circ}$), STOP\}. The episode terminates upon the agent executing the STOP action or after reaching the maximum step limit of 1,000 actions.
The goal of the agent is to leverage the scene information from the prior video to make efficient navigation to the target location with minimal action steps.

\subsection{Instruction Design}
\label{sec:task_design}

\begin{figure}[t]
    \centering
    \includegraphics[width=0.97\linewidth]{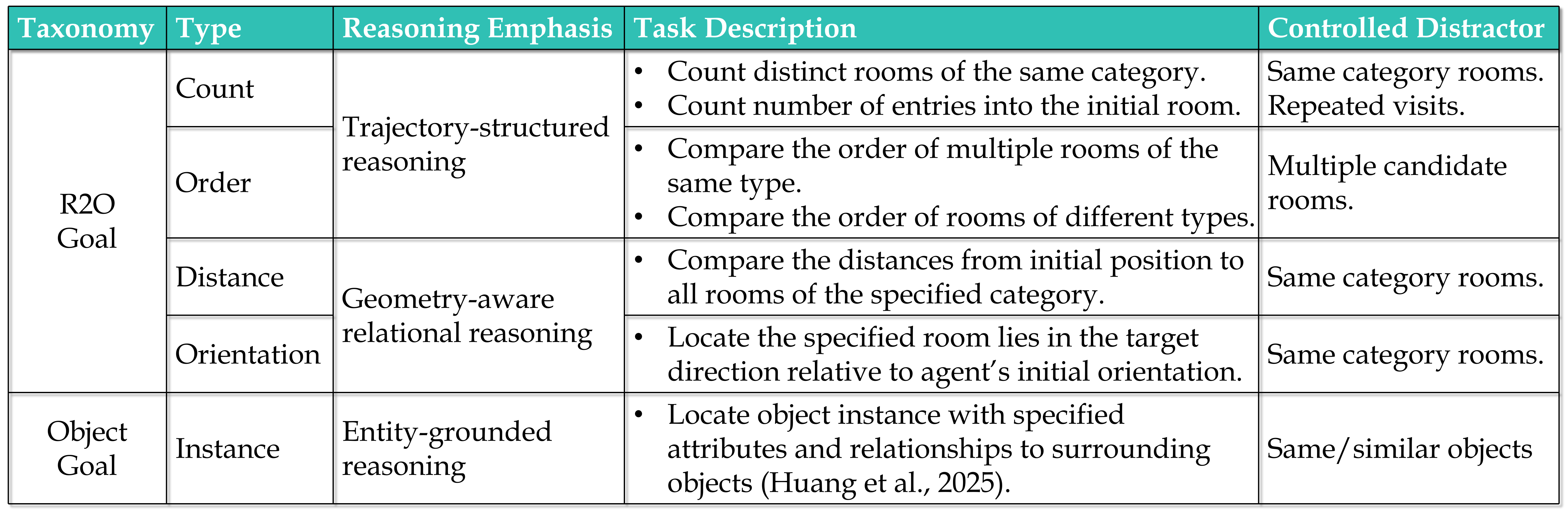}
    \caption{Instruction descriptions and primary dimensions of reasoning over prior visual experience.}
    \label{fig:task_design}
\end{figure}

As shown in \Cref{fig:task_design}, we design five types of instructions that can be classified into two high-level categories: Room-to-Object (R2O) Goal and Object Goal.
\textbf{R2O goal} instructions require reasoning over the rooms visited in the video to identify the target room through Count, Order, Distance, or Orientation relations before referring to an object within it. %, placing different demands on spatial memorization and relative-spatial reasoning.
\textbf{Object goal} instructions require the localization of an object Instance observed in the prior video without extra environmental exploration. %The target objects are guaranteed to be visible in the prior video. 
% \todo
% We present task descriptions and their targeted spatial capabilities for goal identification in \Cref{fig:task_design}, with their samples shown in \Cref{fig:instr}. During navigation, all tasks require full-stack spatial capabilities — iteratively memorizing incremental observations, self-localization in the constructed implicit map, and navigation planning.

The five instruction types emphasize three complementary dimensions of reasoning over prior visual experience. (i) Entity-grounded reasoning, targeting specific objects represented in the prior video. (ii) Trajectory-structured reasoning, focusing on the visitation counts and order of regions along the video trajectory. (iii) Geometry-aware relational reasoning, distinguishing candidate regions according to their distance or relative direction.

\textbf{Design Principles.}
(i) Each instruction is grounded in its prior video. The target and all information required to distinguish it must be covered in the prior video.
(ii) The target cannot be specified (R2O goal) or is hard to specify (Object goal) without the prior video.
(iii) To reduce shortcuts based on category matching or semantic retrieval, we introduce controlled distractors, such as multiple rooms of the same category.
(iv) For R2O goal instructions, the target room can be unambiguously determined from the prior video and the agent’s initial state. For Object goal instructions, we apply instructions~\citep{huang2025vlln} that can uniquely determine the target object.

\subsection{Benchmark Construction}
\label{sec:bench_construct}

\begin{figure}[t]
    \centering
    \includegraphics[width=0.97\linewidth]{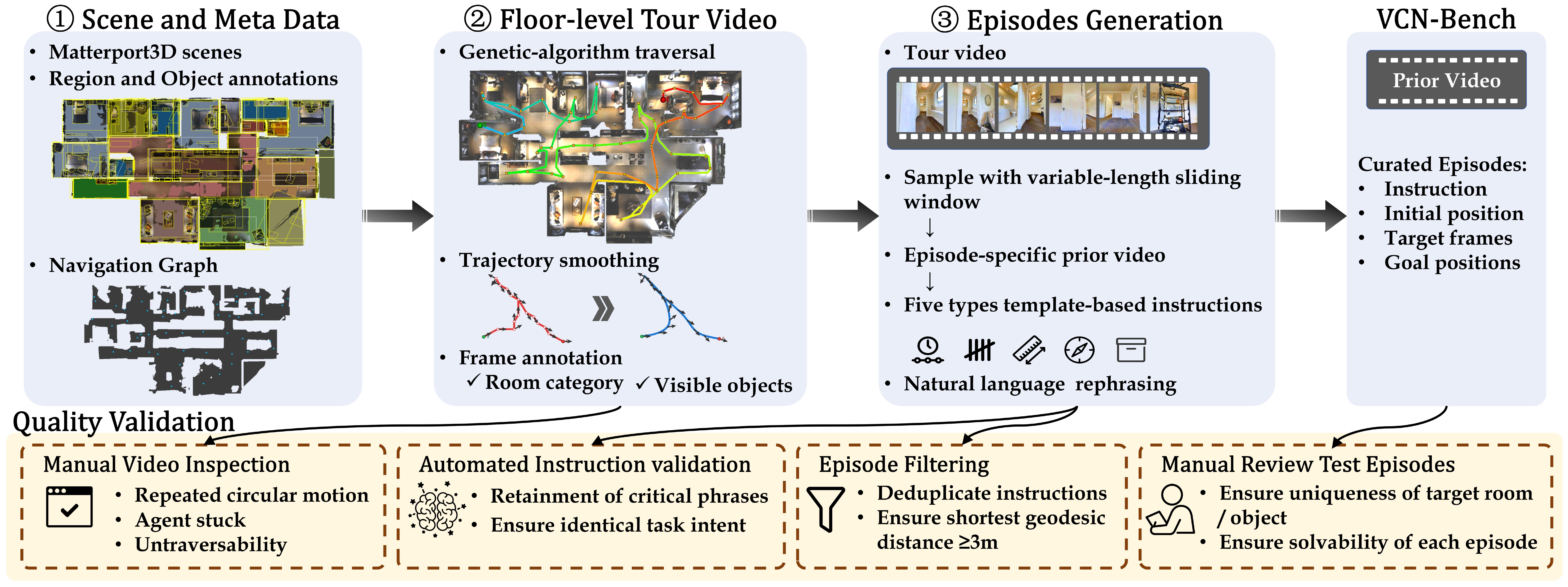}
    \caption{Benchmark construction pipeline.}
    \label{fig:pipeline}
\end{figure}

We apply diverse real-world 3D scans from Matterport3D (MP3D)~\citep{chang2017matterport3d}, paired with region-level annotations. For object-level annotations, we adopt the fine-grained labels from EmbodiedScan~\citep{wang2024embodiedscan}. 
As shown in \Cref{fig:pipeline}, we first collect floor-level tour videos, then sample prior videos from them as episode-specific video context, upon which the corresponding instructions are grounded.

\textbf{Tour Video Generation.}
% To generate a video that thoroughly captures the entire environment, 
For each floor in MP3D scenes, we use the annotated navigation graphs~\citep{anderson2018vision} to compute a minimum-cost traversal sequence covering all navigable nodes via a genetic algorithm~\citep{holland1992genetic}.
We then generate a smooth trajectory using cubic spline interpolation for position and spherical linear interpolation for rotation, to ensure visual continuity. We collect RGB observations along the trajectory to form the floor-level tour video, with randomized camera parameters (height, field of view, pitch angle) and lighting conditions to improve diversity.
For each tour video, we annotate the room category and visible objects per frame. An object is deemed visible if it lies within 3m of the camera and over 60\% of the object can be projected onto the frame without occlusion. See details in \Cref{app:obj_vis}.

\textbf{Episodes Generation.}
% \label{sec:episode}
We construct individual episodes by sampling variable-length sliding windows from the full ``tour video'' to form the ``prior video''. 
Based on the prior video, we construct five types of instructions as detailed in \Cref{sec:task_design} via predefined templates. To make the instructions more natural and open-ended, they are rephrased using Gemini-3.1 Flash.
Since overlapping prior videos may yield identical instructions, we remove duplicate ones to ensure task diversity. Besides, we filter out episodes with a shortest geodesic distance below 3m.

% For each prior video, we annotate the room category and visible objects in each frame, where an object is considered visible if it is within 3m of the camera and more than 60\% of its point cloud is projected onto the frame. We filter out episodes with a shortest geodesic distance below 3m, and remove duplicate episodes to ensure task diversity.

% \textbf{Instructions Generation.}
% We generate navigation instructions via predefined templates, then rephrase them using Gemini-3.1 Flash to ensure natural, open-ended language. We construct 6 types of instructions, divided into two high-level categories, as detailed in \Cref{sec:task_design}.

% deduplication

\textbf{Data Validation and Quality Control.}
To ensure data quality, we conduct human and automated validation.
For tour video generation, we manually inspected all the generated videos to avoid simulator-related failures, such as repeated circular motion. Videos with such failures are regenerated or removed.
During instruction rephrasing, the language model may alter task-critical semantics. We implement a two-level automated validation process to ensure the maintenance of task intent and critical phrases.
For each evaluation episode, we conduct manual validation to ensure unambiguity.
% See more details about benchmark construction and data validation in \Cref{app:benchmark}.

\subsection{Dataset Statistics}

Following the split in MP3D, we employ 61 scenes for training and 11 disjoint unseen scenes for testing.
For each floor within training scenes, up to 5 tour videos are collected, starting from different rooms. Resulting in a training set comprising 309 tour videos and 103,055 episodes.
For each floor in test scenes, one tour video is collected, yielding a test set of 21 tour videos and 1,250 episodes, 250 episodes for each instruction type. 

Test set statistics are shown in \Cref{fig:data}. 
% The entire area of any region visited in the video is included
% We report the number of frames in the prior video, the number of rooms covered, and the area (the entire area of any region visited in the video is included). For each episode, the length of the instruction, the shortest geodesic distance to reach the goal, and the number of steps required for the shortest path are computed.
The memory size of concurrent navigation models~\citep{cheng2024navila,internnav2025,wei2025dulevln} is limited (8-32 frames), while our prior video has an average of 157.9 frames. The average number of steps for VLN-CE~\citep{krantz2020beyond} is 55.8, whereas that of our benchmark is 122.1, more than twice that of VLN-CE.
These statistics suggest that VCN-Bench poses non-trivial challenges for both video understanding and navigation planning.

\section{Baseline}
\label{sec:baseline}
VCN-Bench expects the model to be equipped with three core capabilities: long-term spatial memory retention, spatial reasoning over prior video content, and goal-directed navigation planning. To fulfill these requirements, we propose MV-DualVLN, an MLLM-based navigation model built upon the System 2 of DualVLN~\citep{wei2025dulevln}.
% MV-DualVLN predicts the next navigation waypoint in pixel-space and applies an oracle executor to move to the waypoint. This setting focuses on navigation planning and factors out low-level control failures.

% uses only RGB input, eliminating the need for depth sensors or odometry signals.

\textbf{Preliminary.}
DualVLN synergizes high-level reasoning and low-level control through a dual-system design: System 2 performs navigation planning by predicting turning decisions or mid-term waypoints in image pixel space, while System 1 executes low-level execution by generating continuous trajectories. 
Given the navigation history of 8 frames, current single-view observation, and the instruction, System 2 determines whether to adjust view or output a pixel goal.
% To rigorously evaluate the model's spatial intelligence in navigation tasks and mitigate the interference of low-level control challenges (\eg, obstacle avoidance, robot stuck and fall, \etc) on performance, we build MV-DualVLN based on System 2. To obviate low-level view adjustment, we extend single-view observations to multi-view. The generated pixel-goal is executed by the oracle agent in the simulator. 

\subsection{Model Overview}
% VSNav is built upon Qwen3-VL-4B-Instruct, a strong open-source MLLM with demonstrated capabilities in video understanding and pixel-level spatial grounding. Following previous works~\citep{internnav2025,wei2025dulevln}, we formulate navigation planning as a pixel-goal grounding problem, where the model predicts 2D coordinates in the image as the next navigation waypoint.
Following DualVLN, MV-DualVLN predicts the next navigation waypoint in pixel-space. We apply an oracle executor to move to the predicted waypoint. This setting focuses on navigation planning and factors out low-level control failures (\eg, obstacle avoidance, robot stuck and fall, \etc). Moreover, to obviate low-level view adjustment, we extend single-view observations to multi-view.

\begin{figure}[t]
    \centering
    \includegraphics[width=0.8\linewidth, height=0.23\linewidth, keepaspectratio=false]{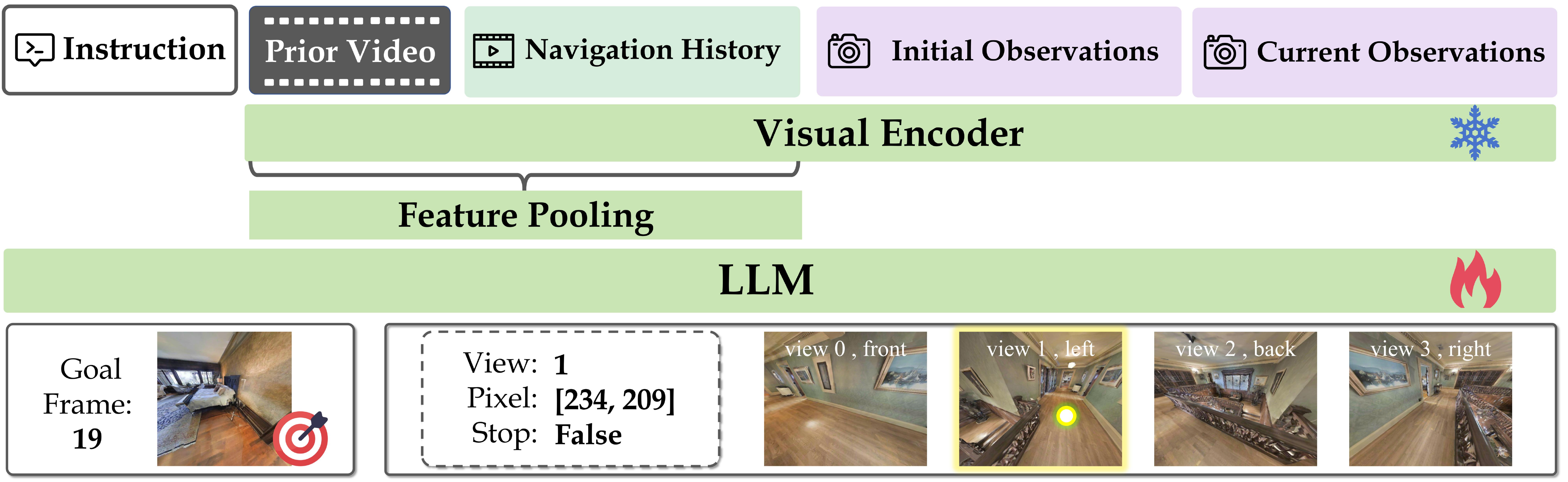}
    \caption{The overall framework of MV-DualVLN.}
    \label{fig:model}
\end{figure}

MV-DualVLN is fine-tuned on Qwen3-VL-4B-Instruct with visual encoder frozen. As illustrated in \Cref{fig:model}, the model inputs consist of: (1) the instruction, (2) the prior video downsampled to 50 frames, (3) in-episode navigation history, (4) multi-view (front/left/back/right) observations at the initial position, and (5) current multi-view observations.
MV-DualVLN performs feature pooling on both the prior video and navigation history to enable long memory retention.
It first identifies the goal by selecting a target frame in the prior video.
Leveraging spatial context jointly encoded from the prior video and navigation history, it then performs navigation planning by predicting—within a selected egocentric observation—a pixel-goal and a termination signal.
% 
% design advantage: 走的时候对环境理解越来越好？
% The merit of this design is that VSNav performs reasoning using RGB input alone, eliminating the need for depth sensors or odometry signals.
% For pixel-goal navigation, we first project the predicted pixel coordinate into a global 3D point in the simulator’s world frame, and then deploy an oracle agent to navigate to that point.

\subsection{Design Details}
\textbf{Navigation History.}
We represent view selection as forward-view rotation, and record the agent’s observations corresponding to rotational actions along with forward actions in the navigation history. For instance, when the agent chooses the right view, the navigation history will encompass the agent's front and right views. This design ensures visual continuity between history frames and integrates additional scene observations from diverse views into the prior video.
% helps the agent perform self-relocalization in the scene and integrate additional scene observations from diverse views into the prior video.

\textbf{Feature Pooling.}
To address the long input sequence challenge, we standardize all input images to $384 \times 384$ resolution, and apply average pooling to downsample the $12 \times 12$ feature map of each image in the prior video and navigation history to $3 \times 3$, reducing the token count per image by $16\times$ while preserving core spatial information. This design enables the model to process long prior videos and navigation history within an acceptable token budget.
% Given that the input includes a substantial quantity of images, we carry out pooling on the image features of the prior video and navigation history to decrease the number of tokens. 
% Specifically, the resolution of all input images is standardized to $384 \times 384$. Thus the length of the feature sequence generated by each image is 144. Given a token budget of 3072, the maximum number of images is only 21, far from the demands of VIS-NavBench tasks.
% To address this, we apply average pooling to the features associated with the prior video and navigation history, downsampling the $12 \times 12$ feature map to $3 \times 3$. The deepstack visual features are pooled identically.

\textbf{Training Data Collection.}
The ground-truth pixel goals for training samples are derived from frontiers. During data collection, an occupancy map of the visited regions is maintained for frontier detection. At each step, the frontier closest to the target location is selected as the next goal waypoint. 
% candidate frontiers are detected in accordance with the maintained occupancy map. We select the frontier closest to the target as the next goal waypoint. 
% We generate ground-truth pixel goals from frontier points detected in the occupancy map, selecting the frontier closest to the target as the next waypoint. 
% This frontier is then mapped to the pixel space of the view which, among the four views of the current observation, has the smallest rotation angle with the frontier.
This frontier is then mapped to the pixel coordinates of the current observation’s most aligned view—that is, the view (among the four available) exhibiting the smallest rotation angle relative to the frontier direction.  
When the target object is included in the occupancy map, and the geodesic distance from the agent's position is less than 3 meters, the termination signal is set to true. At this point, the pixel goal is the pixel-coordinate projection of the object’s centroid.
% the mapping of the object's center within the pixel space.

\section{Evaluation Protocols and Baselines}
\label{sec:eval}

We evaluate MLLMs on VCN-Bench from two perspectives. 
Video-contextualized navigation is the primary evaluation, while goal identification provides a diagnostic to determine whether navigation failures arise from inaccurate destination resolution or from deficiencies in closed-loop interaction.
This section defines both evaluation protocols and metrics, and specifies their baseline methods.
Implementation details are presented in \Cref{app:implement_nav,app:implement_gid}.
% Video-contextualized navigation is the primary evaluation task, measuring whether an agent can use a previously observed scene video and incremental egocentric observations to reach an instruction-specified destination. Since navigation failure alone does not indicate whether an agent identifies the wrong destination or fails during subsequent interaction, we additionally introduce goal identification as a diagnostic protocol. Goal identification removes closed-loop navigation and evaluates whether the model can resolve the intended destination from the prior video.

\subsection{Primary Task: Video-Contextualized Navigation}

\textbf{Protocol.}
% Navigation measures whether an agent can make its way toward the instruction-specified destination from the video through an iterative perception-reasoning-action loop. At each step, it relates its current egocentric observation and navigation history to the prior video, determines the next navigation waypoint, and updates subsequent decisions based on the resulting observation.
Navigation serves as the primary behavioral probe, evaluating the model's capacity to leverage information from the prior video to guide sequential decisions toward the resolved destination amid incremental and changing egocentric observations.

\textbf{Evaluation Metrics.}
We adopt standard evaluation metrics: Success Rate (\textbf{SR}) and Success weighted by Path Length (\textbf{SPL}). An episode is considered successful if the agent stops within 1m Euclidean distance from any valid viewpoint associated with the target.
Additionally, we introduce a relaxed metric: Room Success Rate (\textbf{RSR}), where the episode is deemed successful if the agent ends up in the target room.
The agents are expected to complete navigation by leveraging spatial information embedded in the prior video. Therefore, we adopt \textbf{SR} as the primary evaluation metric. 

\subsection{Navigation Baselines}
We establish two reasoning paradigms. %: hierarchical and end-to-end. 
\textbf{Hierarchical methods} first identify a target frame and then pass it to a separate navigation model. We instantiate this paradigm with a goal-identification model (\Cref{sec:gid_baseline}) followed by MTU3D~\citep{zhu2025mtu}.
\textbf{End-to-End methods} instead use the prior video together with online navigation observations to jointly resolve the destination and make navigation decisions. In addition to our proposed MV-DualVLN, we also evaluate 3D-Mem~\citep{yang20253dmem} and Uni-Navid~\citep{zhang2024uninavid}, both of which are MLLM-driven navigation models capable of goal-oriented navigation and reasoning without depth or camera pose information.

\subsection{Diagnostic Task: Goal Identification}
\textbf{Protocol.}
The model receives the prior video, the task instruction, and multi-view observations at the episode's initial position. It is asked to select the frame in the prior video that best corresponds to the destination specified by the instruction. This protocol performs no navigation actions. 
The selected frame serves as the grounded spatial decision whose correctness depends on the relational structure specified by the instruction.

\textbf{Evaluation Metrics.}
We introduce two metrics: room identification success rate (\textbf{Rid-SR}) for R2O-goal instructions, where success is defined as selecting a frame belonging to the target room; and object identification success rate (\textbf{Oid-SR}) for Object-goal instructions, which is considered correct if the target object is visible in the frame (\Cref{app:obj_vis}).
% within 3m of the selected frame and more than 60\% of its point cloud can be projected onto the frame.

\subsection{Goal Identification Baselines}
\label{sec:gid_baseline}
We evaluate proprietary and open-source MLLMs that support long visual contexts. All models receive the same instruction format and are prompted to return a single frame index.
For proprietary models, we evaluate Claude-Sonnet-5, Qwen3.8-Flash, and 
two variants of Gemini-3-Flash under two inference types. The Nothinking variant directly predicts the target frame, while the Thinking variant enables extended reasoning before producing the frame prediction.
As for open-source models, we include InternVL3.5-8B-Flash~\cite{wang2025internvl35}, GLM-4.6V-Flash, DeepSeek-v4.1-Flash~\citep{xu2026deepseek41f}, and the Qwen3-VL family (4/8B-Instruct/Thinking)~\citep{bai2025qwen3vl}.
\section{Results and Analysis}
\def \ly #1{\cellcolor[HTML]{efefef}{#1}}
\def \lb #1{\cellcolor[HTML]{ECFFFE}{#1}}
\def \lg #1{\cellcolor[HTML]{F0FFF0}{#1}}
\def \lr #1{\cellcolor[HTML]{FFF8F7}{#1}}

\begin{table}[t]
\caption{Comparison of navigation performance. $\dagger$: UniNavid directly outputs low-level actions. Other models utilize an oracle executor to move to the predicted waypoint.}
\label{tab:nav_main}
\setlength{\tabcolsep}{3.8mm}
\resizebox{\textwidth}{!}{%
\begin{tabular}{l|l|c|cccc|cc}
\toprule
&&& \multicolumn{4}{c|}{\textbf{Room-to-Object}} & \multicolumn{1}{c}{\textbf{Object}} \\
\multirow{-2}{*}{\textbf{Models}} & \multirow{-2}{*}{\textbf{Metrics}} & \multirow{-2}{*}{\textbf{Average}} & \textbf{Count} & \textbf{Order} & \textbf{Distance} & \textbf{Orientation} & \textbf{Instance} \\%& \textbf{Category} \\
% \textbf{Methods} & \textbf{Metrics} & \textbf{Average} & \textbf{Count} & \textbf{Order} & \textbf{Distance} & \textbf{Orientation} & \textbf{Instance} & \textbf{Category} \\
\hline
\hline
\multicolumn{8}{l}{\cellcolor[HTML]{ECF4FF}\textbf{\small Hierarchical methods}} \\
\hline
 & \ly{SPL} & \ly{4.8} & {3.4} & {3.4} & {6.2} & {3.3} & {7.9} \\%& {5.1} \\
 & \lb{SR} & \lb{17.6} & {15.6} & {18.4} & {18.4} & {8.8} & {26.8} \\%& {26.4} \\
\multirow{-3}{*}{Gemini-3+MTU3D} 
& \lr{RSR} & \lr{20.5} & {18.8} & {21.1} & {20.0} & {9.6} & {33.2} \\%& {30.4} \\
\hline
 & \ly{SPL} & \ly{5.3} & {5.3} & {4.7} & {5.0} & {3.7} & {7.6} \\%& {7.1} \\
 & \lb{SR} & \lb{20.2} & {22.8} & {22.4} & {18.8} & {11.6} & {25.6} \\%& {32.0} \\
\multirow{-3}{*}{Oracle+MTU3D} 
& \lr{RSR} & \lr{22.0} & {22.8} & {24.0} & {19.2} & {13.2} & {30.8} \\%& {31.6} \\

\hline
\hline
\multicolumn{8}{l}{\cellcolor[HTML]{ECF4FF}\textbf{\small End-to-End methods (zero-shot)}} \\
\hline
 & \ly{SPL} & \ly{2.1} & {2.3} & {1.6} & {2.0} & {1.2} & {3.2} \\%& {2.1} \\
 & \lb{SR} & \lb{5.6} & {8.0} & {4.0} & {4.0} & {5.6} & {6.4} \\%& {4.8} \\
\multirow{-3}{*}{3D-Mem (8B)} & \lr{RSR} & \lr{9.9} & {16.8} & {8.0} & {10.0} & {6.0} & {8.8} \\%& {10.0} \\
\hline
%  & \ly{SPL} & \ly{6.0} & {6.2} & {6.1} & {4.9} & {6.4} & {6.3} \\%& {6.6} \\
%  & \lb{SR} & \lb{8.6} & {8.8} & {7.6} & {6.4} & {8.4} & {12.0} \\%& {10.8} \\
% \multirow{-3}{*}{UniNavid$\dagger$ (w/o video)} & \lr{RSR} & \lr{12.8} & {14.4} & {12.0} & {8.0} & {11.2} & {18.4} \\%& {14.0} \\
% \hline
 & \ly{SPL} & \ly{6.1} & {6.2} & {6.1} & {6.2} & {3.9} & {8.3} \\%& {7.4} \\
 & \lb{SR} & \lb{8.7} & {9.6} & {8.4} & {8.8} & {5.2} & {11.6} \\%& {12.4} \\
\multirow{-3}{*}{UniNavid$\dagger$} & \lr{RSR} & \lr{13.1} & {14.0} & {11.6} & {12.8} & {9.6} & {17.6} \\%& {15.6} \\

\hline
\hline
\multicolumn{8}{l}{\cellcolor[HTML]{ECF4FF}\textbf{\small End-to-End methods (fine-tuned)}} \\
\hline
 & \ly{SPL} & \ly{13.3} & {15.7} & {17.5} & {9.9} & {7.8} & {15.4} \\%& {14.8} \\
 & \lb{SR} & \lb{18.1} & {19.6} & {23.6} & {14.0} & {12.0} & {21.2} \\%& {21.6} \\
\multirow{-3}{*}{3D-Mem (4B)} & \lr{RSR} & \lr{24.6} & {26.4} & {34.4} & {21.2} & {14.8} & {26.4} \\%& {25.2} \\
\hline
 & \ly{SPL} & \ly{19.1} & {22.0} & {23.5} & {14.1} & {11.9} & {23.7} \\%& {20.1} \\
 & \lb{SR} & \lb{27.6} & {30.8} & {34.0} & {20.8} & {18.0} & {34.4} \\%& {30.0} \\
\multirow{-3}{*}{MV-DualVLN (4B)} & \lr{RSR} & \lr{33.9} & {38.0} & {40.4} & {28.2} & {20.4} & {42.4} \\%& {32.0} \\
 \bottomrule
\end{tabular}%
}
\end{table}

This section presents evaluation results and in-depth analyses of both the primary navigation task and the diagnostic goal identification task. More details are presented in \Cref{app:eval} and \Cref{app:ana}.

\subsection{Navigation Performance}

\textbf{Overall performance.} 
\Cref{tab:nav_main} presents the overall navigation performance and task-wise results on VCN-Bench. Our benchmark presents a substantial challenge to all evaluated methods, even the best-performing MV-DualVLN, with an SR of 27.6\%, succeeding in fewer than one-third of navigation episodes.
Overall, current models still struggle to convert scene-specific visual experience into effective sequential navigation decisions.

\textbf{Hierarchical methods.}
MTU3D is provided with two target frame sources: (i) frames selected by Gemini-3-Flash-Thinking, yielding 17.6\% SR; and (ii) ground-truth frames (denoted as Oracle), achieving 20.2\% SR.
We conjecture that underperformance stems from two limitations of MTU3D: its lack of access to spatial context from the prior video and its dependence on CLIP's global features for target representation, hindering fine-grained discrimination among same-category rooms or objects.
% Besides, the marginal gain from Oracle suggests that destination resolution is not the main bottleneck in this pipeline.

% Moreover, substituting GT frames with Gemini-3-selected frames resulted in a slight degradation in navigation performance (SR = 19.1\%), indicating that the performance bottleneck of the hierarchical paradigm lies in action execution.

\textbf{End-to-End methods.}
% Zero-shot 3D-Mem, which employs Qwen3-VL-8B-Instruct for planning, achieves poor results. 
Zero-shot methods yield poor results unsurprisingly.
Fine-tuning 3D-Mem with Qwen3-VL-4B-Instruct on the VCN-Bench improves SR to 18.1\%, demonstrating the importance of task-specific MLLM adaptation in closed-loop tasks.
MV-DualVLN further improves over the fine-tuned 3D-Mem, achieving an absolute SR improvement of 9.5\%, validating the effectiveness of the overall design of our MV-DualVLN.
% rationality of both the pixel-goal paradigm and the navigation history design.

% For Uni-Navid, we evaluate performance with and without the prior video. 
% The limited impact of prepending the prior video is because it is exclusively trained on videos of its own navigation history—data intended for tracking progress rather than supporting scene context.

\textbf{Performance across instruction types.}
The task-wise results reveal systematic differences among destination specifications. For MV-DualVLN, SR is higher on Instance (34.4\%), Order (34\%), and Count (30.8\%) than on Distance (20.8\%) and Orientation (18\%). A similar tendency is observed for the fine-tuned 3D-Mem.
These results suggest that distinguishing candidate rooms according to their geometric relationships with the initial state is particularly challenging for the evaluated methods.

\subsection{Diagnostic Goal Identification Results}

\textbf{Overall performance.}
\Cref{tab:gid} reports the diagnostic goal-identification results.
Although goal identification removes closed-loop interactions, all evaluated models remain substantially below human performance.
These results show that resolving an instruction-specified destination from a long prior video is itself challenging, before the model is required to act in the environment.

\textbf{Chance level performance.}
Complete Chance Level is the success rate of randomly sampling a frame from the prior video.
% Semantic Chance Level is the success rate obtained by randomly sampling a frame from the subset of frames that correspond to the right room category for R2O-goal instructions, or contain an object of the target category for Object-goal instructions.
Category Chance Level is a category-based retrieval shortcut. For R2O-goal tasks, it randomly selects from frames inside rooms matching the target room category; for Object-goal tasks, from frames containing objects matching the target category.
Its underperformance indicates that category-level semantic retrieval alone is insufficient to identify the destination.

\textbf{Human performance.}
We randomly sample a tiny subset of 100 episodes (20 per task type) for human evaluation. Evaluators select a frame using the prior video, initial observations, and instruction. Not surprisingly, humans outperform Gemini-3-flash-thinking by 44.6\% on average Rid-SR and 55.0\% on Oid-SR.
However, they still feel challenged to compare room distances and orientations without a top-down map—relying only on the prior video.

\def \lp #1{\cellcolor[HTML]{ECF4FF}{#1}}

\begin{table}[]
\caption{Comparison of goal identification performance. %$\dagger$: The API has image constraints, prior videos are therefore downsampled to 40 frames. Other models process the full prior video.
}
\label{tab:gid}
\setlength{\tabcolsep}{3.8mm}
\resizebox{\textwidth}{!}{%
\begin{tabular}{l|ccccc|c}
\toprule
 & \multicolumn{5}{c|}{\textbf{Room-to-Object (Rid-SR)}} & \multicolumn{1}{c}{\textbf{Object (Oid-SR)}} \\
\multirow{-2}{*}{\textbf{Models}} & \textbf{Count} & \textbf{Order} & \textbf{Distance} & \textbf{Orientation} & \lp{\textbf{Avg.}} & \textbf{Instance} \\% & \textbf{Category} & \lp{\textbf{Avg.}} \\
\hline
\hline
\rowcolor[HTML]{ECF4FF} \multicolumn{7}{l}{\textbf{\small Baseline}} \\
\hline
Complete Chance Level & 2.0 & 2.4 & 5.6 & 4.0 & \lp{3.5} & 0.8 \\%& 4.4 & \lp{2.6} \\
Category Chance Level & 13.2 & 14.4 & 12.8 & 17.2 & \lp{14.4} & 14.0 \\
\hline
\hline
\rowcolor[HTML]{ECF4FF} \multicolumn{7}{l}{\textbf{\small Tiny Set}} \\
\hline
Human Performance & 93.3 & 95.0 & 86.7 & 83.3 & \lp{89.6} & 95.0 \\
Gemini-3-flash-thinking & 55.0 & 50.0 & 45.0 & 30.0 & \lp{45.0} & 40.0 \\%& 20.0 & \lp{25.0}\\
% GPT-5-High$\dagger$ & 45.0 & 35.0 & 30.0 & 15.0 & \lp{31.3} & 20.0 \\
\hline
\hline
\rowcolor[HTML]{ECF4FF} \multicolumn{7}{l}{\textbf{\small Proprietary Models (API)}} \\
\hline
Gemini-3-flash-nothinking & 52.4 & 40.0 & 46.4 & 28.8 & \lp{41.9} & 36.0 \\%& 24.8 & \lp{30.4} \\
Gemini-3-flash-thinking & \textbf{56.0} & {50.4} & {53.6} & 34.8 & \lp{{48.7}} & {43.6} \\%& \textbf{32.4} & \lp{\textbf{38.0}} \\
Claude-Sonnet-5 & 51.6 & 50.0 & 42.8 & 40.8 & \lp{46.3} & 38.0 \\
Qwen3.8-Flash & 54.0 & \textbf{56.8} & \textbf{60.0} & \textbf{46.0} & \lp{\textbf{54.2}} & \textbf{47.6} \\
\hline
\hline
\rowcolor[HTML]{ECF4FF} \multicolumn{7}{l}{\textbf{\small Open-source Models}} \\
\hline
InternVL3\_5-8B-Flash & 26.0 & 20.8 & 17.6 & 19.2 & \lp{20.9} & 19.2 \\
GLM-4.6V-Flash & 18.8 & 13.2 & 18.4 & 18.8 & \lp{17.3} & 20.0 \\

Qwen3-VL-4B-Instruct & 25.2 & 24.4 & 23.6 & 16.0 & \lp{22.3} & 24.0 \\%& 17.6 & \lp{20.8} \\
Qwen3-VL-4B-Thinking & 31.2 & 21.6 & 25.2 & 21.6 & \lp{24.9} & 20.8 \\%& 20.0 & \lp{20.4} \\
Qwen3-VL-8B-Instruct & 34.9 & 22.3 & 30.0 & 32.5 & \lp{30.0} & 33.5 \\%& 23.1 & \lp{28.3} \\
Qwen3-VL-8B-Thinking & 45.6 & 38.0 & 40.0 & \textbf{41.6} & \lp{41.3} & 32.8 \\%& 27.6 & \lp{30.2} \\
DeepSeek-v4.1-Flash & \textbf{50.8} & \textbf{50.8} & \textbf{48.0} & 40.8 & \lp{\textbf{47.6}} & \textbf{50.4} \\
% \hline

% \rowcolor[HTML]{ECF4FF} \multicolumn{1}{l|}{\textbf{\small SFT Models}} &  &  &  &  &  &  &  &  \\
% Qwen3vl (50frames) &  &  &  &  & \lp{} &  &  & \lp{} \\
\bottomrule
\end{tabular}%
}
\end{table}

\textbf{Comparison across tasks.}
Across R2O-goal tasks, models generally perform better on Count and Order than on Distance and Orientation. This pattern indicates that the evaluated models are more capable of understanding the visitation sequence of the prior video than geometry-aware destination specifications.
Object-goal tasks, requiring fine-grained grounding of a unique instance, is more difficult than R2O-goal tasks for most of the evaluated models.

% Overall, locating target object is more demanding than identifying target room from the video, with average Rid-SR across all models—including the chance-level baseline—consistently lower than average Oid-SR.
% Across all R2O tasks, models generally exhibit superior performance on temporal tasks (Count and Order) relative to spatial tasks (Distance and Orientation). This disparity arises because spatial tasks impose an additional cognitive demand: the implicit construction and maintenance of a coherent cognitive map. The systematic underperformance reflects that existing MLLMs have limitations in mental map construction. %Furthermore, closed-source models outperform open-source counterparts significantly on temporal-reasoning tasks.

% It is surprising to find that Qwen3-VL-8B demonstrates exceptional performance on Orientation tasks. The Instruct variant outperforms Gemini-3-flash-nothinking (32.5\% VS. 28.8\%), while the Thinking variant even surpasses Gemini-3-flash-thinking (41.6\% VS. 34.8\%). This indicates that Qwen3-VL-8B possesses strong capabilities in understanding camera orientation and the global spatial layout of scenes.
% In contrast, on Distance tasks—a fellow spatial task—Qwen3-VL-8B lags significantly behind closed-source models, with the Thinking variant achieving a Rid-SR that is 13.6\% lower than that of Gemini-3-flash-thinking. This finding implies that, despite both tasks requiring reasoning over the cognitive map, estimating distance and orientation tap into distinct capabilities.

\textbf{Comparison across models.}
Thinking mode improves performance across models at the same scale. Gemini-3-flash-thinking achieves absolute improvements of 6.8\% in average Rid-SR and 7.6\% improvement in Oid-SR over its nothinking mode. These gains underscore the critical role of deep reasoning in spatial cognition. 
Scaling model parameters also results in performance gains. For example, Qwen3-VL-8B-Thinking yields an absolute improvement of 16.4\% in average Rid-SR and 12.0\% in Oid-SR over Qwen3-VL-4B-Thinking.

% Among open-source models, scaling up parameter size can yield a significant enhancement in performance.
% Notably, the gap between open-source and proprietary models is not that substantial: Qwen3-VL-8B-Thinking achieves performance comparable to that of Gemini-3-flash-nothinking, with absolute deficits of only 0.6\% in average Rid-SR and 0.2\% in average Oid-SR.

\subsection{Analyses and Ablations}

\begin{figure}
\centering
\begin{minipage}[t]{0.3\textwidth}
    \centering
    \includegraphics[width=\linewidth,height=1.1\linewidth, keepaspectratio=false]{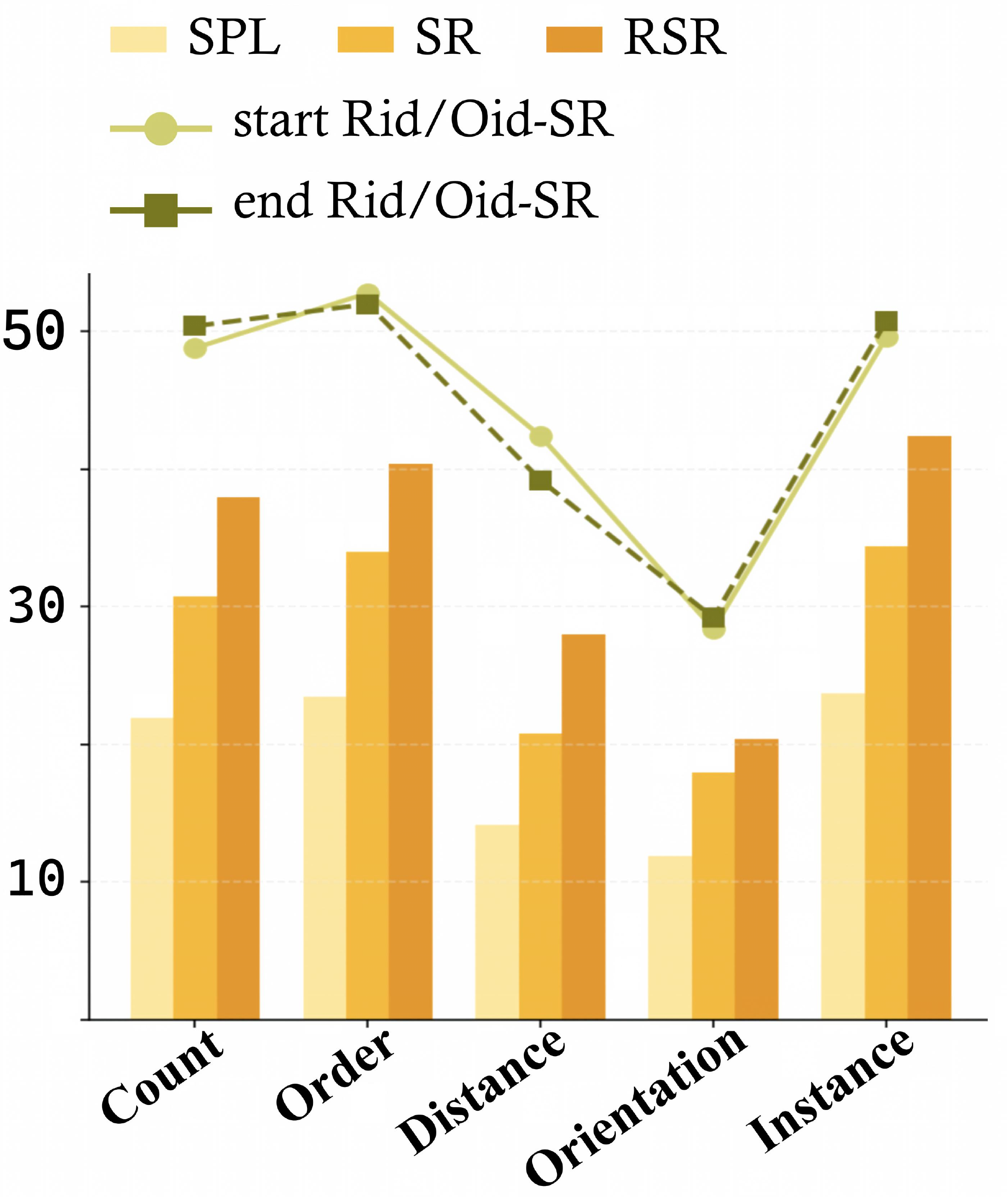}
    \caption{Navigation and Gid results of MV-DualVLN.}
    \label{fig:gid_nav}
\end{minipage}
\hfill
\begin{minipage}[t]{0.65\textwidth}
    \centering
    \includegraphics[width=1\linewidth]{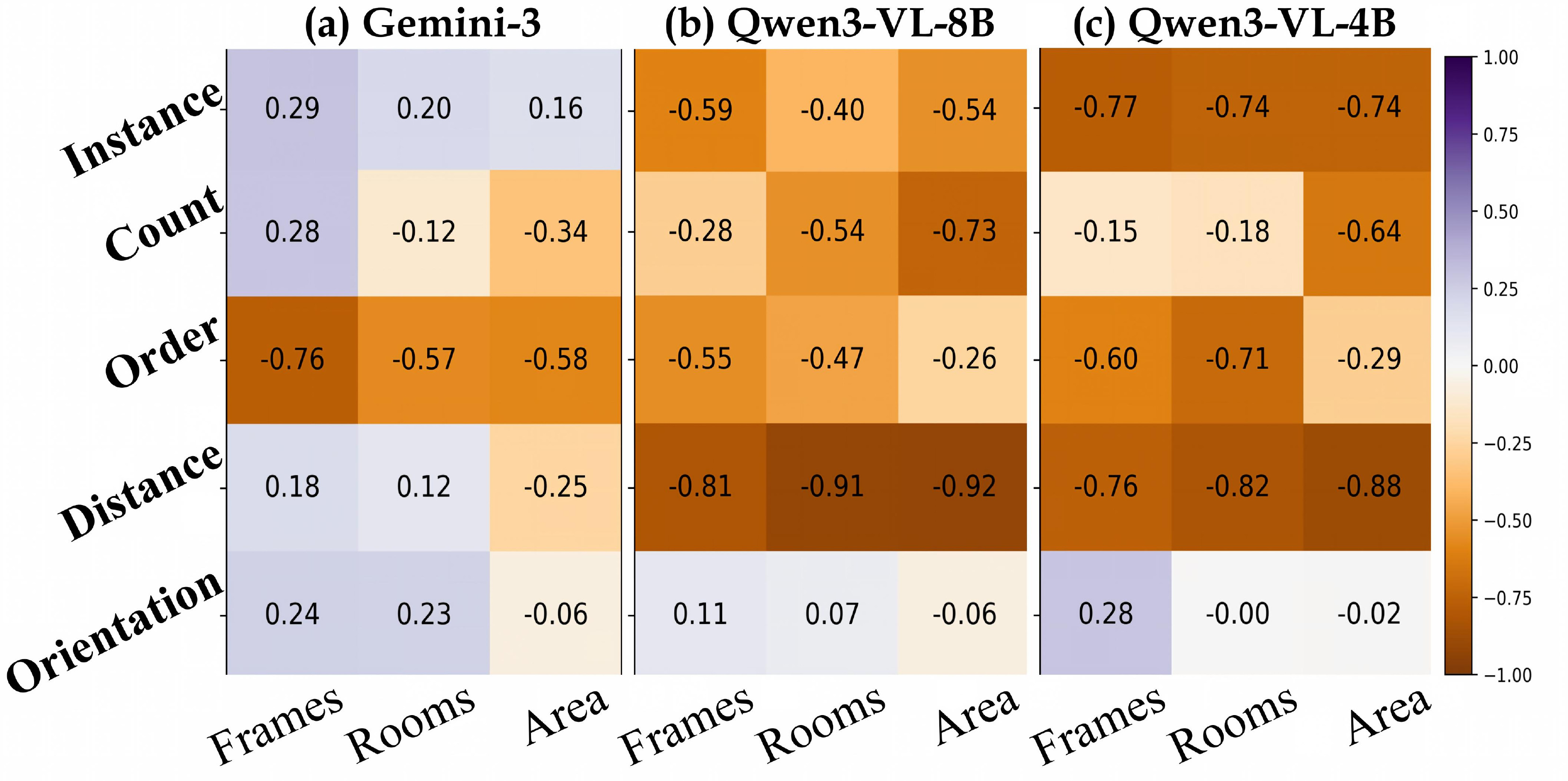}
    \caption{Pearson correlation between model performance and prior video.}
    \label{fig:heatmap}
\end{minipage}
\end{figure}

\textbf{Relationship between goal identification (Gid) and navigation.}
\Cref{fig:gid_nav} illustrates Gid and navigation performance of MV-DualVLN across tasks. Since Gid is performed at every step, we report Gid results at the first step (denoted as start) in light green and the last step (end) in dark green.
A strong positive correlation is evident between Gid and navigation, yet a substantial gap remains.
To describe this gap, we compare Rid-SR and RSR for R2O tasks, both of them are target-room metrics. 
For Object tasks, we compare Oid-SR and a softened SR (SSR), defined as episode success when the agent stops within 3 m of the target object—a comparable distance threshold with Oid-SR’s criterion.
MV-DualVLN achieves an average Rid-SR of 43.1\% but only 31.7\% RSR on R2O tasks, leaving a gap of 11.4 points. The same pattern occurs on Object tasks: Oid-SR reaches 49.6\% 
while SSR is 39.2\%, resulting in a 10.4-point gap.
% while SR is 34.4\%, resulting in a 15.2-point gap.

Moreover, we analyze successful navigation $\mathbf{N}$ conditioned on correct goal identification at the first step $\mathbf{G}$. MV-DualVLN achieves 37.8\% SR among episodes when Gid is correct, \ie, $P(\mathbf{N} \mid \mathbf{G})=37.8\%$, compared to $P(\mathbf{N} \mid \neg \mathbf{G})=18.6\%$.
For R2O tasks, RSR achieves 49.5\% among episodes when Gid is correct, versus 18.1\% RSR when Gid fails.
These results demonstrate that goal identification is an important prerequisite for reliable navigation, yet even when the destination is correctly identified, 62.2\% of episodes still fail to reach the target.
% These results demonstrate that accurate goal identification is a necessary—but insufficient—prerequisite for successful navigation. Even when the target destination is correctly identified, models fail in 62.2\% of episodes to leverage this information to guide subsequent navigation decisions within the closed-loop process.

% \def \ly #1{\cellcolor[HTML]{FFFFF2}{#1}}
\def \ly #1{\cellcolor[HTML]{efefef}{#1}}
\def \lb #1{\cellcolor[HTML]{ECFFFE}{#1}}
\def \lg #1{\cellcolor[HTML]{F0FFF0}{#1}}
\def \lr #1{\cellcolor[HTML]{FFF8F7}{#1}}

% \begin{table}[]
% \caption{Ablation results for navigation. Both variants are trained under the ablation setting.}
% \vspace{-5pt}
% \label{tab:nav_abl}
% \setlength{\tabcolsep}{3.0mm}
% \resizebox{\textwidth}{!}{%
% \begin{tabular}{l|ll|c|cccc|cc}
% \toprule
% \# & \textbf{Variants} & \textbf{Metrics} & \textbf{Average} & \textbf{Count} & \textbf{Order} & \textbf{Distance} & \textbf{Orientation} & \textbf{Instance} & \textbf{Category} \\
% \hline
%  && \ly{SPL} & \ly{16.7} & \ly{17.9} & \ly{22.1} & \ly{11.8} & \ly{10.5} & \ly{19.8} & \ly{18.0}  \\
%  && \lb{SR} & \lb{25.3} & \lb{26.0} & \lb{33.6} & \lb{18.4} & \lb{18.4} & \lb{28.8} & \lb{26.4}  \\
% \multirow{-3}{*}{1} & \multirow{-3}{*}{100 frames, w/o history} 
% & \lr{RSR} & \lr{33.7} & \lr{38.0} & \lr{46.4} & \lr{28.0} & \lr{22.8} & \lr{38.4} & \lr{28.8} \\
% \hline
%  && \ly{SPL} & \ly{12.2} & \ly{11.5} & \ly{14.2} & \ly{10.4} & \ly{4.8} & \ly{17.7} & \ly{14.8}  \\
%  && \lb{SR} & \lb{19.7} & \lb{17.6} & \lb{22.4} & \lb{16.8} & \lb{6.4} & \lb{28.8} & \lb{25.2}  \\
% \multirow{-3}{*}{2} & \multirow{-3}{*}{w/o prior video} 
% & \lr{RSR} & \lr{24.8} & \lr{22.8} & \lr{24.8} & \lr{21.2} & \lr{10.8} & \lr{38.0} & \lr{30.0} \\
%  \bottomrule
% \end{tabular}%
% }
% \vspace{-18pt}
% \end{table}

\begin{wraptable}{r}{6.3cm}
\centering
\vspace{-13pt}
\caption{Ablation results for navigation. Variants are trained under the ablation setting.}
\label{tab:nav_abl}
\resizebox{1.0\linewidth}{!}{
\begin{tabular}{ll|ccc}
\toprule
\# & Models & SPL & SR & RSR \\
\midrule
1 & MV-DualVLN & 19.1 & 27.6 & 33.9 \\
\midrule
2 & target frame only & 12.2 & 19.7 & 24.8 \\
3 & 50 frames, w/o history & 14.9 & 24.1 & 32.8 \\
4 & 100 frames, w/o history & 16.4 & 25.0 & 33.7 \\
\bottomrule
\end{tabular}
}
\vspace{-12pt}
\end{wraptable}

\textbf{Benefits from the prior video.}
To validate that the models indeed acquire scene knowledge of the environment from the prior video, and thus benefit navigation, we present the performance when only the ground-truth target frame is provided rather than the prior video in \Cref{tab:nav_abl} row \#2. 
This variant utilizes the ground-truth target frame as a hint during both the training and testing phases.
It can be observed that performance declines significantly when only a target frame is provided rather than the complete prior video (6.9\% drop in SPL, 7.9\% drop in SR, and 9.1\% drop in RSR).
This indicates that MV-DualVLN indeed leverages the scene context in the prior video throughout navigation.

\textbf{Necessity of navigation history.}
Navigation history records the agent's episodic trajectory and supplements scene information. To validate its effectiveness, we present ablation results in \Cref{tab:nav_abl} rows \#3 and \#4. 
When the prior video is downsampled to 50 frames as MV-DualVLN does, excluding in-episode history results in a 3.5\% decrease in SR and 4.2\% in SPL.
Under a token budget comparable to MV-DualVLN, downsampling prior video to 100 frames without navigation history achieves on-par RSR with MV-DualVLN, yet SR remains 2.6\% lower.
The comparison between rows \#3 and \#4 indicates that higher video frame density is beneficial for spatial reasoning.
Comparing MV-DualVLN with row \#4 reveals that incorporating in-episode navigation history compensates for scene observation and enhances object localization within the target room.  

% To ensure fairness, we set the token budget at 2048, allocate 200 tokens for text, and subtract observations at the initial and current positions. The remaining tokens can accommodate 77 pooled images.
% MV-DualVLN takes 50 frames of prior video, setting aside 27 frames for navigation history.
% In row \#4, prior video is downsampled to 100 frames with no navigation history, slightly exceeding the budget. 
% It can be observed that retaining navigation history yields better performance, with the SPL and SR being 2.5\% and 2.7\% higher, but the RSR being 0.2\% lower.
% This indicates that navigation history is beneficial for finding objects in the target room.

\textbf{Relationship between model performance and prior video.}
% \setlength{\columnsep}{3pt}
% \begin{wrapfigure}{r}{0.55\textwidth}  % r=右侧放图，0.4\textwidth=图宽
%   \vspace{-15pt}
%   \setlength{\columnsep}{0pt}
%   \centering
%   \includegraphics[width=1\linewidth]{figures/heatmap.png}  % 替换你的图片
%   \vspace{-16pt}
%   \caption{Pearson correlation between model performance and prior video.}
%   \vspace{-12pt}
%   \label{fig:heatmap}
% \end{wrapfigure}
% Taking Gemini-3-flash-thinking and Qwen3-VL-4/8B-Thinking as representative models, we quantify the relationships between their performance and three key attributes of the prior video—namely the number of frames, the number of rooms covered, and the area covered—using the Pearson correlation coefficient.
We quantify the correlation between performance of three representative models and three key attributes of the prior video, \ie, number of frames, number of rooms covered, and covered area, using the Pearson correlation coefficient.
\Cref{fig:heatmap} presents the correlation heatmap. % illustrating the associations between each of these three models and the prior video across six task categories.
It can be observed that Gemini-3 and Qwen3-VL models exhibit different correlations. For Gemini-3, only Order tasks show a negative correlation with frame count, indicating robustness to video length and spatial coverage.
In contrast, both Qwen3-VL variants show consistent negative correlations across all three video attributes, reflecting sensitivity to longer videos and broader scene coverage.
% , while the correlation in other tasks is relatively weak. This indicates that Gemini-3 is more robust to video length and covered space. 
% In contrast, both Qwen3-VL models exhibit a negative correlation between performance and the three attributes of the prior video, indicating sensitivity to increases in length and coverage space of the input context.
% For all three models, the performance on Orientation tasks exhibits a small correlation with the three attributes of the prior video, suggesting that this task is more related to the model's inherent capabilities.

\section{Conclusion}
In this paper, we propose VCN-Bench, a video-contextualized navigation benchmark for probing closed-loop spatial reasoning over prior visual experience in MLLMs. Built upon diverse indoor scenes, our benchmark provides a video as the scene-specific spatial context of the navigation task. 
VCN-Bench contains five instruction types and uses closed-loop navigation as its primary evaluation, accompanied by diagnostic goal identification to distinguish destination-resolution errors from subsequent navigation failures. We further propose MV-DualVLN as a reference baseline and support in-depth analysis.
Experiments reveal limited navigation performance and a substantial destination-resolution-to-navigation gap: correct destination identification improves navigation success, but many correctly identified episodes still fail during subsequent interaction.
% Despite all the evaluations and analyses, we acknowledge that due to the high cost of proprietary MLLM APIs, coupled with the input constraints of models, we have trouble conducting evaluations across a wider range of proprietary MLLMs.

\bibliographystyle{unsrtnat}
\bibliography{main}

\beginappendix
\clearpage

\section{More Related Works}
\textbf{MLLM-based Navigation Models.}
Recent advances in MLLMs~\citep{yang2024qwen25,Qwen3-VL,guo2025deepseek-r1,hong2025glm} have driven a new wave of embodied navigation models.
% Driven by the rapidly advancing understanding and generalization capabilities of multimodal large language models (MLLMs), a growing body of recent research has leveraged MLLMs for embodied navigation tasks. 
Early explorations frame MLLMs as zero-shot planners~\citep{zhou2023navgpt,chen2024mapgpt,long2024discussnav,qiao2024opennav,zhang2025flexvln,chen2026harnessvln}, while recent Vision-Language-Action (VLA) models~\citep{zheng2024navillm,cheng2024navila,zhang2024uninavid,wei2025streamvln,zeng2025janusvln,zheng2025efficientvln,xue2025omninav,zhang2025navfom,gao2026octonav,wang2026lightnav} finetune MLLMs on large-scale embodied data and demonstrate notable improvements on navigation benchmarks.
However, as constrained by existing tasks, their closed-loop spatial reasoning capability remains underexplored. We evaluate existing MLLM-driven navigation models on our benchmark and propose MV-DualVLN as a reference baseline for VCN-Bench.

\section{Benchmark Details}
\label{app:benchmark}
\subsection{Tour video VS. Prior video}
\begin{figure}[h]
    \centering
    \includegraphics[width=1\linewidth]{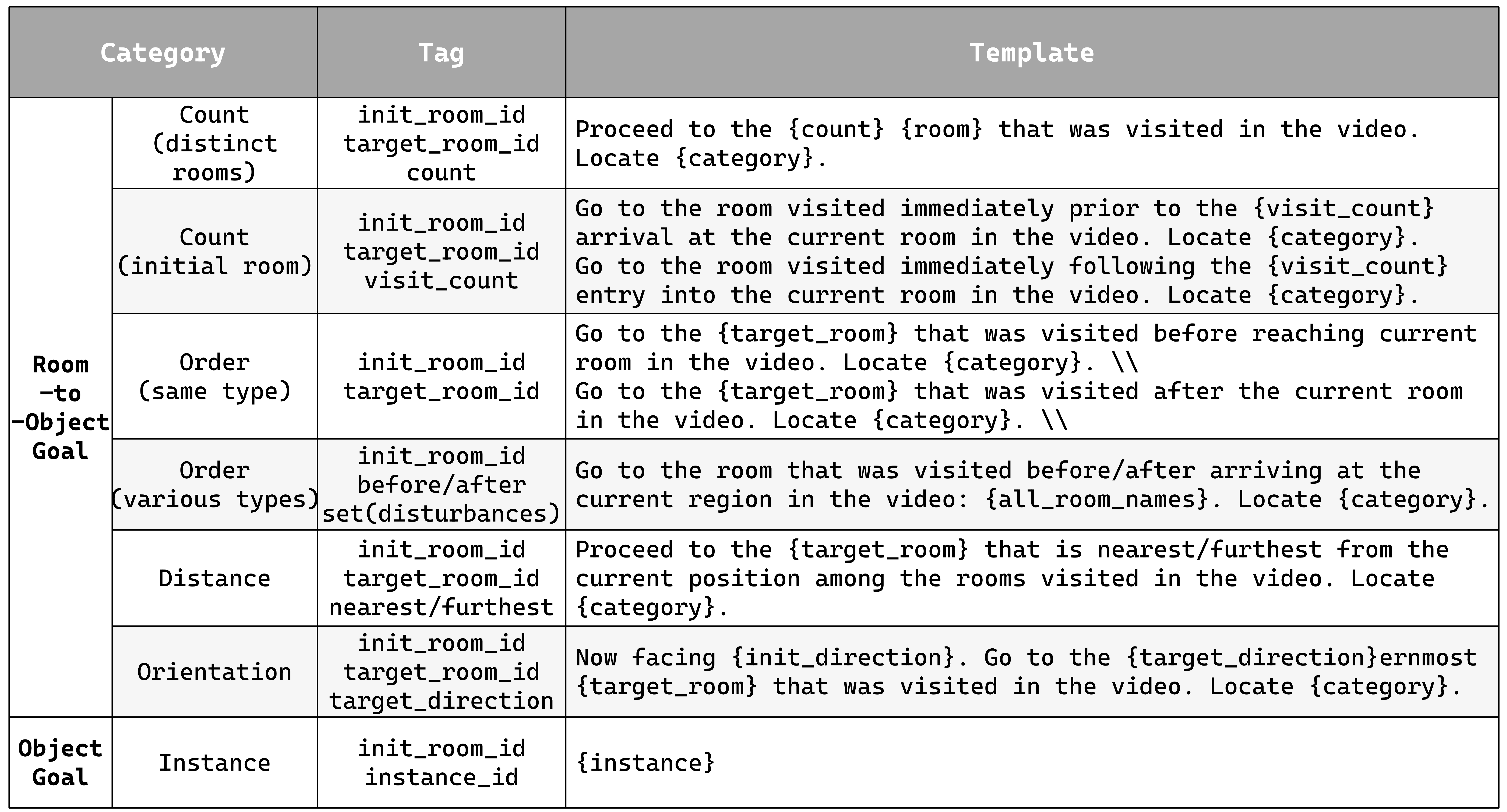}
    \caption{Templates and tags of instructions.}
    \label{fig:app_template}
\end{figure}

A tour video comprehensively covers a floor. Nevertheless, when constructing an episode, it is only necessary to ensure that the agent's initial and target positions fall within the video coverage, and there is no necessity to utilize the complete tour video. Therefore, we employ sliding windows of varying lengths to extract prior videos from the tour video and subsequently construct episodes based on these prior videos. 
Specifically, we adjust the size of the sliding window at intervals of 10 frames. Regarding the starting position of the window, we guarantee that the region corresponding to the starting frame differs from the previous one. Given that there is an overlap between prior videos, different prior videos may yield identical episodes. Therefore, as depicted in \Cref{fig:app_template}, we tag the instructions and eliminate episodes with duplicate tags.

\subsection{Region Annotation}
All the region types in MP3D are as follows:
\begin{figure}[h]
    \centering
    \begin{tcolorbox}
    stairs, closet, bathroom, hallway, workout/gym/exercise, other room, toilet, familyroom/lounge, laundryroom/mudroom, dining room, garage, outdoor, spa/sauna, living room, kitchen, bar, entryway/foyer/lobby, balcony, porch/terrace/deck, office, rec/game, bedroom, lounge
    \end{tcolorbox}
\end{figure}

When constructing R2O instructions, to ensure clarity of the task, the initial room and the target room should avoid areas with ambiguous boundaries, including: hallway, other room, balcony, entryway/foyer/lobby, porch/terrace/deck, outdoor.

When annotating the region for each frame in the tour video, we use the high-quality region annotations provided by Matterport3D itself. However, due to irregular room shapes, 5.8\% of frames fall within multiple rooms simultaneously. We manually determine the correct region using a top-down view that contains both region annotations and frame positions.

\begin{figure}[th]
    \centering
    \includegraphics[width=1\linewidth]{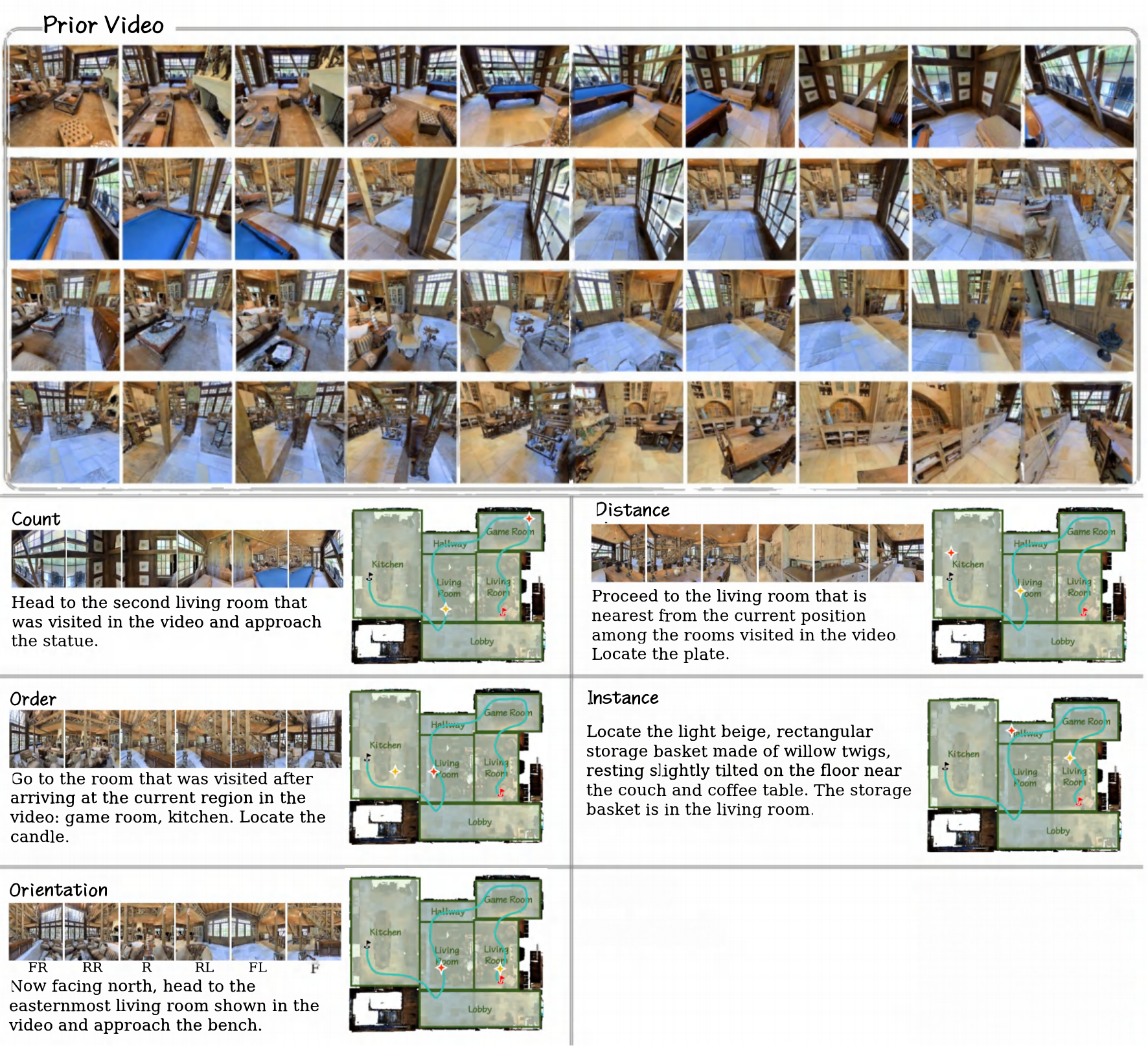}
    \caption{Example of VCN-Bench. In the top-down map, we plot the video trajectory with a green curve, which starts from the red flag and ends at the black flag. The red and yellow stars are the markers for the initialization and target positions of each episode. ``FR'', ``RR'', ``R'', ``RL'', ``FL'', ``F'' are the abbreviations of Front-Right, Rear-Right, Rear, Rear-Left, Front-Left, Front, respectively.}
    \label{fig:app_bench}
\end{figure}

\subsection{Object Visibility Definition}
\label{app:obj_vis}
When evaluating goal identification performance on Object goal tasks, we need to determine whether the target object instance is visible in the selected frame.
An object instance is considered visible in a frame when it lies within 3m of the camera position and more than 60\% of its projected surface can be included in the frame without occlusion.
Specifically, we have access to the bounding box and point cloud of objects on the same floor, along with per-frame camera poses and depth maps. 
We compute the minimum horizontal distance from the camera position to the object's point cloud, ignoring the vertical coordinate, and retaining objects within 3m.
For the remaining objects, we project their point cloud onto the image plane and construct an object-specific depth map through z-buffering, retaining only the nearest object depth at each pixel. The resulting pixels approximate the object's projected front surface under the current camera view.
We compare this object depth map with the rendered scene depth map corresponding to the frame. A front-surface pixel is considered visible if the two depths differ by less than 0.1m.
We require more than 60\% of the object's in-frame front-surface pixels to pass this depth-consistency test.
We additionally filter objects that occupy fewer than $50 \times 50$ visible pixels within the $384 \times 384$ frame resolution.

\subsection{Instruction Design Principles}
The instruction templates are shown in \Cref{fig:app_template}. Tags are employed to eliminate identical instructions generated by prior videos of varying lengths within the same scene.
For R2O goal instructions, the agent must locate an instance of the target object category after reaching the target room. The target room may contain multiple instances of the category. Success is achieved upon reaching any such instance.
% is required to locate the target object within the target room. The target room may contain multiple object instances of the target category, navigating to any of them is considered successful.

\textbf{Order.}
The first subclass involves tracking the order of multiple rooms of different types and the agent's initial room, such as ``Go to the room that was visited before arriving at the current region in the video: bathroom, living room. Locate the cabinet." 
Take this instruction as an example: if the target room is the bathroom, the prior video should meet the conditions: (1) the initial room is only visited once in the video; (2) there is only one bathroom within the video; (3) the bathroom may be visited multiple times prior to the entry into the initial room; and (4) the bathroom will not be revisited after the visit of the initial room.

The second subclass involves comparing the appearance order of multiple rooms of the same type and the initial room, such as ``Go to the bedroom that was visited after reaching the current room in the video. Locate the lamp." 
For this instruction, (1) there should be multiple bedrooms visited in the prior video. We assume there are bedroom1 and bedroom2 visited in the prior video,  and the target room is bedroom2. The prior video should also satisfy: (2) the initial room is only visited once in the video; (3) bedroom2 is the only bedroom visited in the video after leaving the initial room.

\textbf{Count.}
The first subclass involves counting the number of distinct rooms of the same category, such as ``Proceed to the second bedroom that was visited in the video. Locate the shelf."
For this instruction, (1) the prior video must contain multiple bedrooms. We assume there are bedroom1 and bedroom2 visited in the prior video,  and the target room is bedroom2. (2) The visitation of bedrooms should be in a clear order. For example, bedroom1 can be visited multiple times before entering bedroom2, and it will not be revisited after leaving bedroom2.

The second subclass involves counting the number of entries into the agent's initial room, such as ``Go to the room visited immediately following the second entry into the current room in the video. Locate the mirror."
For this type of instruction, (1) the initial room should be visited for at least two times.

\textbf{Distance.}
This category involves comparing the distances between the initial position and all candidate rooms, such as ``Proceed to the bathroom that is farthest from the current position among the rooms visited in the video. Locate the lamp".
Take this instruction as an example: (1) there should be multiple bathrooms visited in the prior video. Assume there are bathroom1 and bathroom2 within the coverage of the prior video, with bathroom1 designated as the target room. (2) To ensure disambiguation, the geodesic distance between the initial position and the center of bathroom1 must be at least 5m larger than that between the initial position and the center of bathroom2.

\textbf{Orientation.}
This category involves identifying the direction of candidate rooms conditioned on the initial rotation.
To formulate these instructions, identifying the regions within the video coverage that are situated in the four directions is a prerequisite.
In the Habitat coordinate system, we define the direction indicated by the Z-axis as south and the direction indicated by the X-axis as east. 
For all the regions accessed in the video, denote their horizontal bounding box as $[x_{min}, x_{max}, z_{min}, z_{max}]$.
Taking the south direction as an example, the maximum value of all $z_{max}$ is defined as $max\_Z$. All rooms that meet the condition: $max\_Z-5 < z_{max} <= max\_Z$ are regarded as candidate rooms. Within the X-axis range of a candidate room, if there is no room with its center point located further south, this room is considered to be at the southernmost part.

Take the instruction ``Now facing north. Go to the easternmost closet that was visited in the video. Locate the carpet." as an example. Within the coverage of the prior video: (1) there should be at least two closets; (2) among all the easternmost rooms, there should be only one closet at the easternmost side; (3) to avoid task ambiguity, the closet cannot be located in the northeast or southeast corner.

\textbf{Instance.}
We employ the instance descriptions in VLLN~\citep{huang2025vlln}, where the unique identification of the target instance is achieved via the object's inherent attributes and its relative positional relationships with surrounding objects.

% \textbf{Category}.
% Taking the instruction "Locate the table seen in the video" as an example, we aim for the agent to identify the table closest to the initial position by leveraging the prior video. To this end, (1) the closest table to the initial position must be observable in the prior video. Nevertheless, the agent is permitted to select an alternative table, so (2) all tables within the video’s coverage area qualify as valid goals.

\subsection{Instruction Rephrasal}
We utilize Gemini-3.1-Flash to rephrase the template-generated instructions to ensure natural and open-ended language. The prompt is as follows:

\begin{figure}[h]
    \begin{tcolorbox}
        Optimize the navigation instruction to enhance naturalness and task-appropriateness.
        The instruction is generated based on a tour video. Do not confuse the video with the agent's own navigation history.
        Rather than uniformly using “locate”, the verb should be semantically aligned with the target object's function or intended interaction—for example, “open the cabinet”, “turn on the TV”.
        Retain \{retain\_info\} information.
        
        Instruction: \{ori\_instr\}
        
        Rephrased Instruction:
    \end{tcolorbox}
\end{figure}

Each instruction type contains certain crucial information that must be retained. For example, for Distance tasks, it is essential to retain the phrase "among the rooms visited in the video" to guarantee the accuracy of the task.

% After rephrasing, we also guarantee the accuracy of the instructions through rules. For example, regarding Count (distinct rooms) instructions, if the target room is the second bedroom visited in the video, then it is necessary to ensure that the rephrased instruction still includes ``second bedroom".

\subsection{Data Validation Details}
To ensure data quality, we conduct human and automated validation.
For tour video generation, we manually inspected all the generated videos to identify simulator-related failures, including prolonged collisions, repeated circular motion, abnormal camera behavior, or trajectories that fail to cover their intended regions. Videos containing such failures are regenerated or removed.

During instruction rephrasing, the language model may alter task-critical semantics.
% introduce unintended alterations to task-critical semantics.
We implement a two-level automated validation process. Firstly, we ensure the task-critical phrases are retained. For example, if the target room is the second bedroom visited in the video, then the ``second bedroom" phrase should be retained. Secondly, we employ Gemini-3-Flash to evaluate whether the original and rephrased instructions convey identical task intent. Instructions failing either level of validation are automatically reverted to the template-based version.

Finally, we manually verify the unambiguity of all evaluation episodes. Two experts are provided with the prior video along with a top-down scene map with video trajectory. Besides, they are also provided with an oracle-generated navigation video showing the optimal path from the episode’s initial position to the destination. 
They assess whether (i) the target room or object can be uniquely determined by the instruction from the prior video, and (ii) the episode is objectively solvable (\ie the destination is reachable).

\subsection{Dataset Statistics}
\begin{figure}[t]
    \centering
    \includegraphics[width=0.97\linewidth]{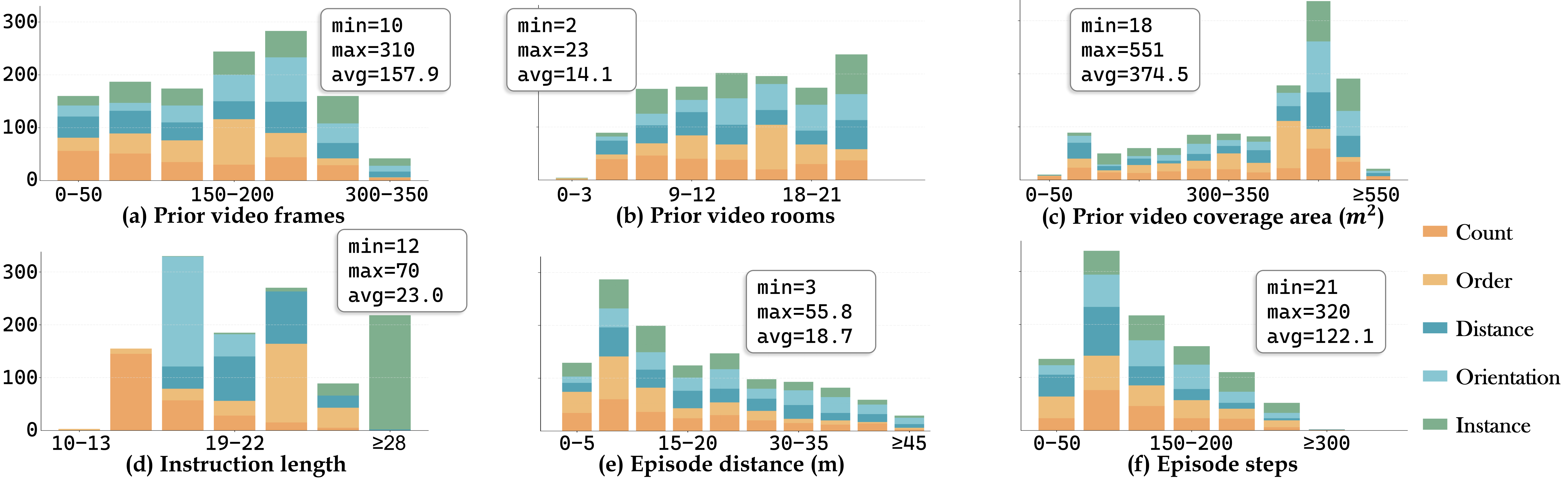}
    \caption{Benchmark statistics, including (a) the number of prior video frames, (b) number of rooms and (c) area (the entire area of any region visited is included) covered by the prior video, (d) number of words in instructions, (e) the shortest geodesic distance, and (f) action steps of episodes.}
    \label{fig:data}
\end{figure}

As illustrated in \Cref{fig:data}, we report statistics of both the prior video (\ie, frame count, number of distinct rooms traversed, and total covered floor area) and the corresponding episodes (\ie, instruction length, shortest navigation distance, and required action step count).

\subsection{Benchmark Examples}
Benchmark samples are presented in \Cref{fig:app_bench}. The prior video that has been downsampled to 40 frames is displayed at the top. Examples of 5 types of instructions with corresponding egocentric observations of the agent at the initial position are shown on the right.
Additionally, we visualize the top-down map of the scene, where the red and black flags represent the starting and end points of the prior video, and the green curve depicts the video trajectory. The initialization position and the destination of each episode are marked with red and yellow flags.

\begin{figure}[th]
    \centering
    \begin{tcolorbox}
        \#\#\# Role\\
        You are a highly capable indoor navigation and visual reasoning assistant. You excel at video understanding and determining navigation targets.\\

        \#\#\# Input Definition\\
        You will be provided with:\\
        1. **Room Tour Video**: A sequence of \{chunk\_size\} frames capturing a spatial layout.\\
        2. **Current Observations**: 4 images illustrating your current position.\\
        3. **Instruction**: A description of the target destination.\\
        
        \#\#\# Your Reasoning Process\\
        1. **Spatial Layout Analysis**: Carefully examine the "Room Tour Video" to understand the spatial layout of the current scene.\\
        2. **Current Position Determination**: Analyze the "Current Observations" to determine your current position within the scene. Consider the orientation of the images and any visual cues that indicate your location.\\
        3. **Target Destination Identification**: In accordance with the given "Instruction", identify the target destination and select the best frame from the "Room Tour Video" that most accurately represents both the destination and the target (in the target room or at the closest distance to observe the target object).\\ %When only the object category is specified in the instruction, the object of the specified category that is closest to the current position shall be located.\\
        
        \#\#\# Output Format\\
        Please respond ONLY with a JSON object containing the following keys:\\
        - "Thought": A brief, step-by-step logical derivation of your current location and the target's frame index.\\
        - "frame\_index": The integer index (0 to \{chunk\_size-1\}) of the target destination frame in the "Room Tour Video".\\
        
        ----\\
        \#\#\# Input Data\\
        - Room Tour Video: 0:\texttt{<image>} 1:\texttt{<image>} 2:\texttt{<image>} ...\\
        - Current Observations: front\texttt{<image>} left\texttt{<image>} back\texttt{<image>} right\texttt{<image>}\\
        - Instruction: \{instr\}
    \end{tcolorbox}
    \caption{Prompt for zero-shot goal identification.}
    \label{fig:prompt}
\end{figure}

\section{Evaluation Details}
\label{app:eval}

\subsection{Navigation Success Definition}
Point clouds are collected at intervals of 0.1m within a range of 0.6m around the bounding box of the target objects. The points located in the same room as the target instance are retained as viewpoints.
If the minimum distance between the agent's final stopping position and all viewpoints is less than 1m, the task is considered successful. In comparison with the traditional object navigation tasks, which require the agent to stop within 1m of the object, our metric is slightly more lenient.
Navigation success is uniformly defined across all models as reaching a position within 1m of the viewpoints.

\subsection{Agent Setup}
For the navigation evaluation, we adopt the Hello Robot Stretch platform, which has a height of 1.41 m and a base radius of 0.17 m.  The action space includes MOVE\_FORWARD (0.25m), TURN\_LEFT (30$\degree$), TURN\_RIGHT (30$\degree$), and STOP. 

\subsection{Implementation Details of Navigation Baselines}
\label{app:implement_nav}

We establish two reasoning paradigms for video-contextualized navigation evaluation. The advantage of the \textbf{Hierarchical paradigm} lies in decoupling goal identification from navigation planning—each task is assigned to a domain-specific expert model. However, it suffers from the limitation that the navigation model cannot exploit environmental structural information embedded in the prior video.
\textbf{End-to-End paradigm} necessitates concurrent goal identification and navigation planning, imposing more stringent demands on models' reasoning capabilities. However, this paradigm enables the model to fully leverage scene information embedded in both the prior video and in-episode navigation history, thereby facilitating efficient navigation.

\subsubsection{MTU3D}
MTU3D represents both detected objects and exploration frontiers as queries, and the model is trained to select a query according to the navigation instruction. The oracle executor then executes low-level actions to reach the selected query.

We apply MTU3D as the navigation expert in the hierarchical reasoning paradigm. Specifically, MTU3D is provided with the instruction and a target frame—either selected by Gemini-3-Flash-Thinking or drawn from ground-truth annotations. 
During evaluation, MTU3D first performs image-goal navigation toward the provided target frame. Since this frame may not be close enough to the target object, MTU3D then performs object-goal navigation, using instance descriptions for Instance tasks and object categories for other task types.

\subsubsection{MV-DualVLN}
During image processing, all images are initially resized to $384 \times 384$. Qwen3-VL divides the image into $16\times 16$ patches and applies a $2\times$ spatial compression in the image encoder, resulting in an overall compression rate of 32. Thus, each image, after being processed by the image encoder, yields a total of $12\times 12$ tokens. We conduct feature pooling on these image features and compress them into $3\times 3$ tokens. Qwen3-VL also utilizes the image features of the intermediate layer; they are compressed in the same manner. Feature pooling can notably reduce the training time.

MV-DualVLN is trained for 5 epochs, with the vision encoder frozen. We adopt a learning rate of $1\times 10^{-5}$ with a cosine learning rate decay scheduler. The model is trained in bfloat16 format using 32 A100 GPUs with a batch size of 2. The training takes about 22 hours.

\subsubsection{3D-Mem}
3D-Mem is a zero-shot navigation framework that represents visited regions and exploration frontiers in the form of RGB images, and applies an MLLM to choose an image as the next navigation subgoal.
Specifically, 3D-Mem performs object detection on all historical observations acquired during navigation, clusters detected objects according to their 3D spatial coordinates, and selects the minimal set of images that collectively encompass all detected objects—these are termed memory snapshots. Concurrently, 3D-Mem maintains an occupancy map to identify exploration frontiers, and designates observations oriented toward such frontiers as frontier snapshots. 

At each navigation step, the MLLM selects a single snapshot from the unified pool of memory and frontier snapshots as the next waypoint. %Selection of a frontier snapshot signifies an intent to expand exploration in that direction, whereas selection of a memory snapshot indicates successful localization of the target object within the snapshot—thereby terminating navigation.
When a frontier snapshot is selected, it signifies the agent intends to move in that direction, and the next waypoint is set at a predefined distance along that direction.
When a memory snapshot is selected, it signifies successful visual localization of the target object within that snapshot’s field of view; the next waypoint is then computed as the 3D centroid of all detected objects in the snapshot, and navigation terminates upon reaching this point. In both cases, the oracle executor is applied to execute low-level actions to reach the intended waypoint.

We evaluate 3D-Mem on VCN-Bench under two settings: (i) zero-shot inference using Qwen3-VL-8B-Instruct as the MLLM, and (ii) fine-tuned 3D-Mem initialized from Qwen3-VL-4B-Instruct. For fine-tuning, we adopt the same training protocol as MV-DualVLN: the prior video is downsampled to 50 frames, and feature pooling is applied jointly over features extracted from both the prior video and the memory snapshots.

\subsubsection{Uni-Navid}
Uni-Navid is a video-based navigation model trained on massive navigation and video QA data. It takes in an RGB video stream and outputs low-level actions.
To adapt it to VCN-Bench, we prepend the prior video to the in-episode navigation history.

\subsection{Implementation Details of Goal Identification}
\label{app:implement_gid}
For consistency, all models are evaluated with a temperature of zero without output length constraints.
The prompt for goal identification is presented in \Cref{fig:prompt}.

\subsection{Human Evaluation}
\textbf{Setup.} 
During the evaluation of human-level goal identification performance on VCN-Bench (tiny), three human evaluators are allowed unlimited time to select target frames to the best of their ability.
They receive the prior video, the instruction, multi-view observations, and the initial position of the episode. They are allowed to review the prior video multiple times to fully understand the scene.
Their averaged performance is reported in \Cref{tab:gid}.

\textbf{Human errors.}
Human errors in goal identification stem from three primary sources:
\begin{itemize}
    \item Due to relying solely on visual input for turns and forward actions in the video, without proprioceptive feedback from physically walking through the environment, leading to errors in judging distance and direction.  
    \item Ambiguity in initial pose localization. Observations are restricted to four static, egocentric views (front, back, left, right) from a fixed position, precluding active exploration to disambiguate room identity. For instance, when multiple visually similar closets appear in the video, and the agent initializes inside one. Without exiting the room to observe, it is easy to misidentify which closet one is currently in.  
    \item Careless selection errors. For instance, selecting a frame where the agent is still at the doorway of the target room but has not entered yet; 
    or overlooking some spatial descriptors (\eg, relative positions of nearby objects) specified in the instance instruction.
\end{itemize}

% \begin{figure}[t]
%     \centering
%     \includegraphics[width=0.7\linewidth]{figures/downsample2.pdf}
%   \caption{Goal identification performance under 3 sampling rates.}
%     \label{fig:downsample}
% \end{figure}
% ,height=0.6\linewidth, keepaspectratio=false

\section{More Analysis}
\label{app:ana}

\begin{figure}[t]
    \centering
    \includegraphics[width=0.9\linewidth]{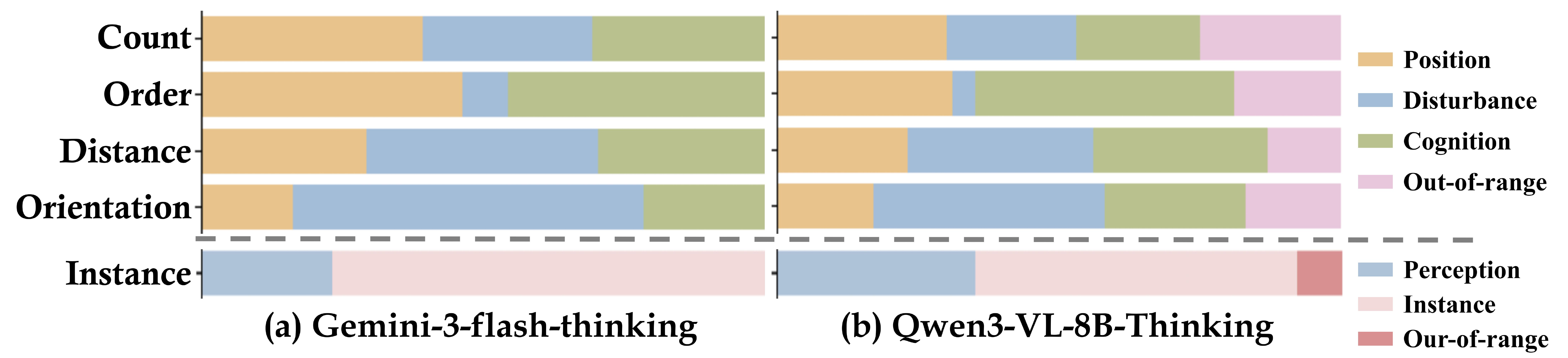}
    \caption{Error analysis of goal identification.}
    \label{fig:mllm_error}
\end{figure}
\begin{figure}[t]
    \centering
    \includegraphics[width=0.7\linewidth]{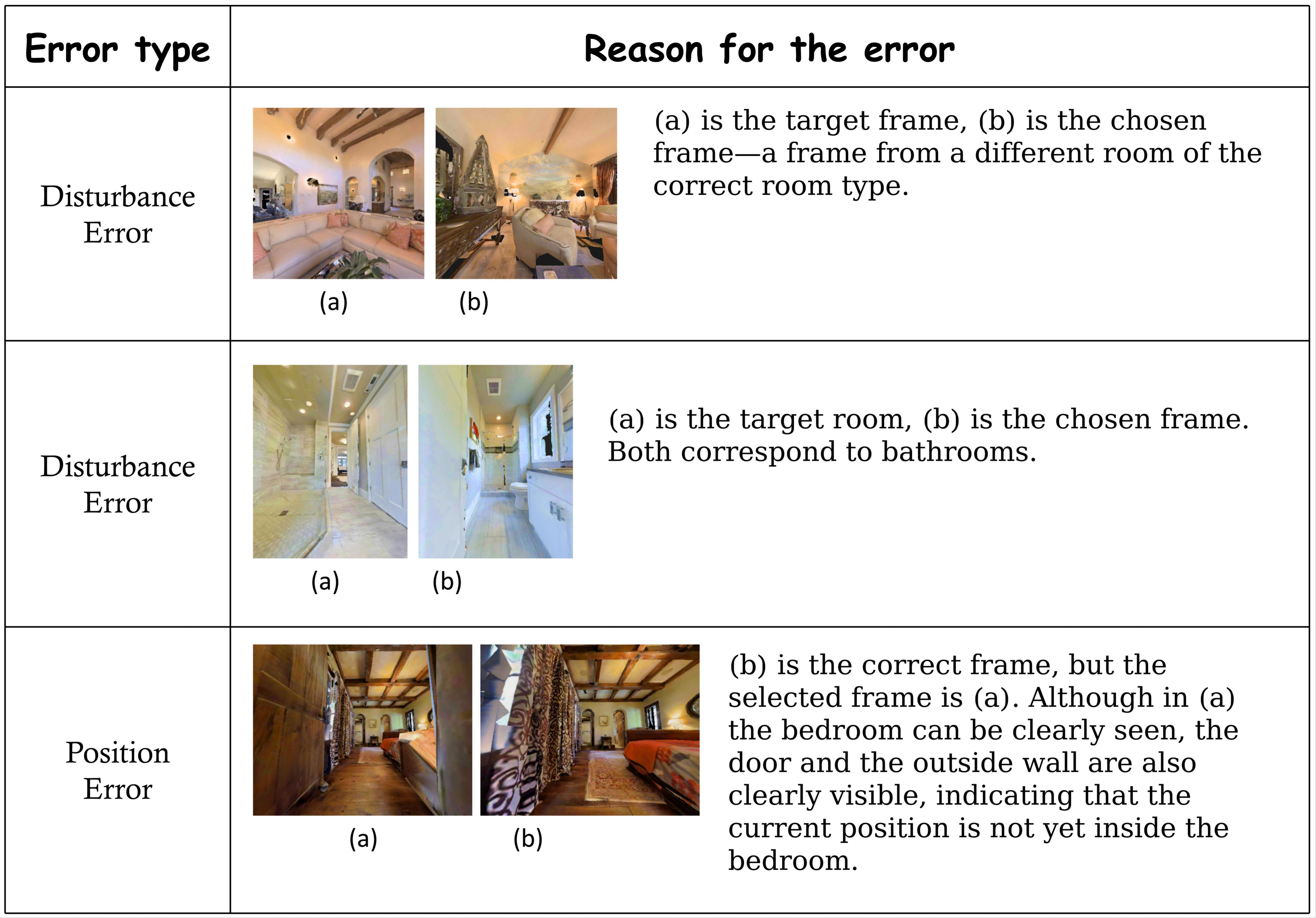}
    \caption{Visualization of Disturbance and Position error.}
    \label{fig:err_vis1}
\end{figure}

\subsection{MLLM Errors Analysis}
To gain a more in-depth understanding of the reasons behind MLLM's errors, we categorize and analyze the causes of errors in goal identification. 
% We select Gemini-3-flash-preview-thinking, which exhibits the best zero-shot performance.

Errors in R2O goal tasks can be classified into four categories: (i) Position error, the selected frame contains the target room but is positioned outside the room. (ii) Disturbance error, the selected frame is positioned in the wrong room of the right room type. If the selected frame contains the disturbance room but is positioned outside the room, it is also classified under this error type. (iii) Cognition error, the selected frame is unrelated to the target room. (iv) Out-of-range error, the predicted frame index is larger than the frame count of the video.

For Object goal tasks, the errors can be classified into three categories: (i) Instance error, the selected frame encompasses an object belonging to the correct category, but it is not the specific object instance that the instruction is intended for. (ii) Perception error, the selected frame does not contain any object of the target category, or the object is located at an excessive distance. (iii) Out-of-range.

As presented in \Cref{fig:mllm_error}, we analyse the goal identification errors of Gemini-3-flash-thinking and Qwen3-VL-8B-Thinking. Our findings are as follows:
\begin{itemize}
    \item Gemini-3 does not exhibit Out-of-range errors, whereas the quantity of this error type in Qwen3-VL is not negligible. This indicates that Gemini-3 demonstrates a more comprehensive understanding of the input context, while Qwen3-VL encounters a considerable number of hallucination issues.
    \item Position errors are common across all R2O tasks. This is because a more comprehensive view of the room can typically be obtained from outside, whereas once inside, the field of vision becomes restricted. This implies that MLLM exhibits a greater reliance on directly observable information and has a relatively limited understanding of its own position and the spatial relationship between frames.
    \item For Distance and Orientation tasks, which necessitate selecting a target room from rooms within the same category, disturbance error is the predominant factor. For Count and Order tasks, which entail making selections among different types of rooms, cognition error is the primary factor. These error distributions are in line with the challenges associated with the tasks.
    \item Perception error is the primary error for Instance tasks. This may be because the instructions of instance tasks involve numerous relationships between the target object and other surrounding objects, leading the model to misidentify the surrounding objects as the target ones.
    % \item For Category tasks, there are multiple objects of the target category within the scene. We expect the model to select the object closest to the current position, which is also clearly stated in the prompt~\Cref{fig:prompt}. The main cause of error in Category tasks is Instance error, suggesting that the model failed to select the nearest target object.
\end{itemize}

\begin{figure}[t]
    \centering
    \includegraphics[width=1\linewidth]{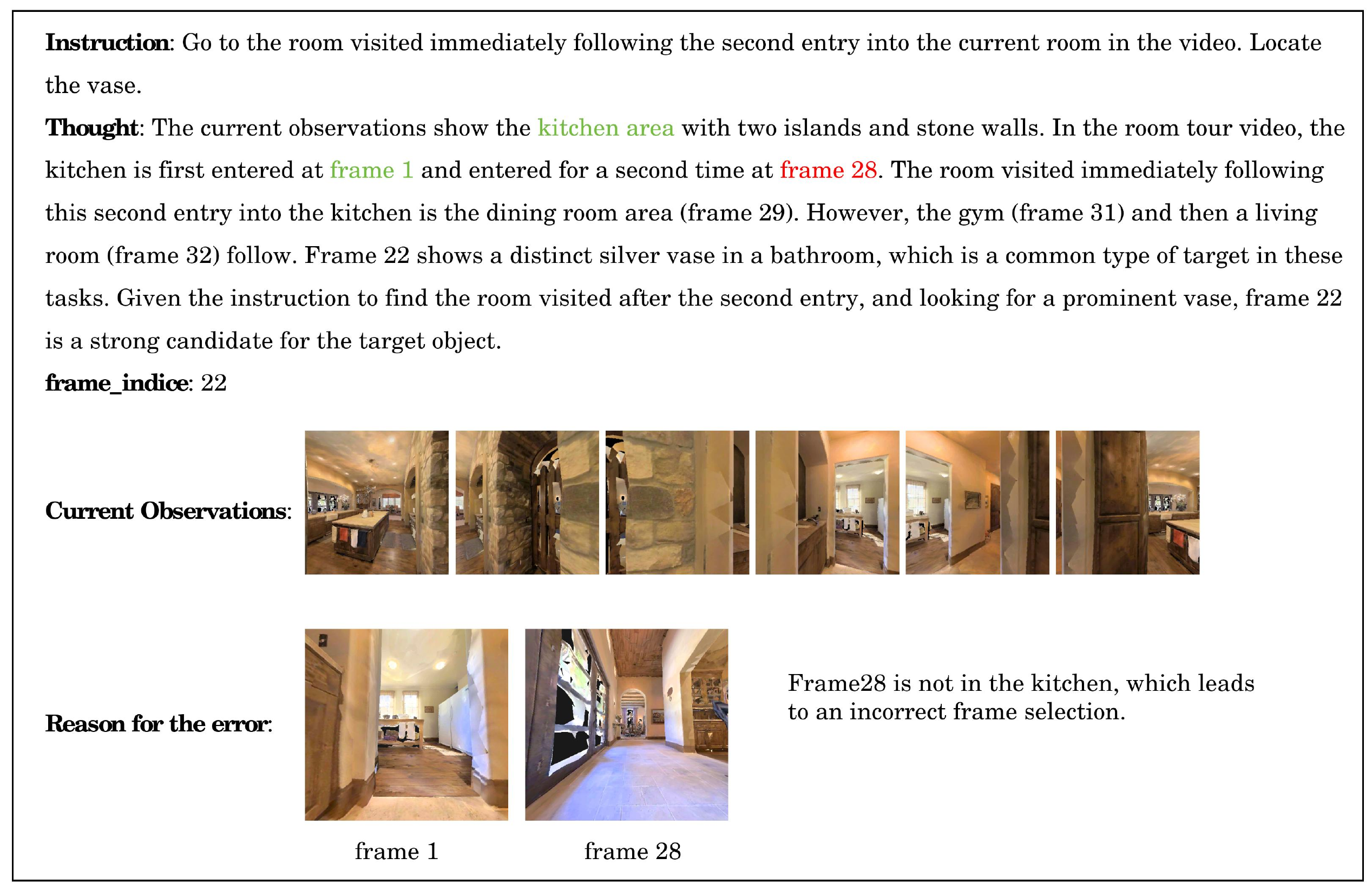}
    \caption{Visualization of Cognition error.}
    \label{fig:err_vis2}
\end{figure}

We visualize cases for Disturbance error and Position error in \Cref{fig:err_vis1}. The case for Cognition error is depicted in \Cref{fig:err_vis2}.

\subsection{Analysis on Spatial Context Utilization}
In this section, we conduct three controlled experiments to investigate the capacity of MV-DualVLN in leveraging spatial context.
All experiments use the same trained MV-DualVLN model and the same evaluation protocol without additional fine-tuning. The interventions modify only the information available to the model during inference, allowing us to probe destination utilization, relational sensitivity and video order sensitivity.

\begin{wraptable}{r}{6cm}
\centering
% \caption{Performance comparison on navigation when}
% \label{tab:nav_abl}
% \setlength{\tabcolsep}{1mm}
% 
\vspace{-12pt}
\resizebox{1.0\linewidth}{!}{
\begin{tabular}{l|ccc}
\toprule
Models & SPL & SR & RSR \\
\midrule
MV-DualVLN & 19.1 & 27.6 & 33.9 \\
% \midrule
GT frame prediction & 22.8 & 33.1 & 39.9 \\
\bottomrule
\end{tabular}
}
\vspace{-12pt}
\end{wraptable}
\paragraph{Navigation with Oracle target frame.}
In the standard MV-DualVLN setting, the model first identifies a target frame from the prior video and then predicts the next navigation waypoint. Thus, the waypoint decision explicitly conditions on the predicted target frame. This motivates a question: how would navigation performance change if the target frame predictions were perfect?

Therefore, we replace the model-predicted target frame with the ground-truth target frame and provide it as the goal-frame prefix to MV-DualVLN throughout navigation. All other model inputs, visual processing, and navigation decisions remain unchanged.
As shown in the table on the right, conditioning on ground-truth target frames yields nontrivial improvements in navigation performance: SPL increases by 3.7\%, SR by 5.5\%, and RSR by 6.0\%.
This suggests that correct target resolution can partially rectify erroneous navigation decisions.  

\paragraph{Counterfactual Relation Intervention.}
We conduct a counterfactual intervention to examine whether MV-DualVLN is sensitive to the relational semantics specified in the instruction. For each evaluation episode, we keep the prior video, initial state, scene, and original navigation target unchanged, while replacing a task-critical spatial or temporal relation in the instruction with its contradictory counterpart. The interventions for each R2O goal instruction type are as follows:
\begin{itemize}
    \item Count: The ordinal in the instruction (first, second, \etc) is replaced by a different one.
    \item Order: The temporal relation is reversed by swapping ``before'' and ``after''.
    \item Distance: The distance relation is reversed by swapping ``nearest'' and ``farthest''.
    \item Orientation: The agent’s initial orientation is maintained, while the direction of the target room (northernmost, southernmost, easternmost, or westernmost) is replaced by a different one.
\end{itemize}

As shown in \Cref{tab:internvention}, intervention leads to substantial performance degradation across all tasks. Particularly, SR dropped by 27.6\% on Order tasks. This provides empirical evidence that MVDualVLN relies critically on the relational semantics in the instructions for reasoning.

\begin{table}[h]
\centering
\setlength{\tabcolsep}{3.8mm}
\caption{Navigation performance comparison on instruction types with interventions.}
\label{tab:internvention}
\resizebox{\textwidth}{!}{%
\begin{tabular}{l|ccc|ccc|ccc}
\toprule
      & \multicolumn{3}{c|}{MV-DualVLN} & \multicolumn{3}{c|}{Counterfactual Relation} & \multicolumn{3}{c}{Disordered Video} \\
\cline{2-10}
      & SPL & SR & RSR & SPL & SR & RSR & SPL & SR & RSR \\
\midrule
Count & 22.0 & 30.8 & 38.0 & 7.1 & 9.9 & 15.0 & 8.8 & 10.8 & 16.8 \\
Order & 23.5 & 34.0 & 40.4 & 4.2 & 6.4 & 10.8 & 7.3 & 10.0 & 15.6 \\
Distance & 14.1 & 20.8 & 28.2 & 6.5 & 9.2 & 11.6 & 7.8 & 11.6 & 17.2 \\
Orientation & 11.9 & 18.0 & 20.4 & 4.4 & 7.2 & 13.2 & 6.6 & 9.2 & 14.8 \\
Instance & 23.7 & 34.4 & 42.4 & - & - & - & 11.9 & 16.8 & 26.0 \\
\bottomrule
\end{tabular}
}
\end{table}

\paragraph{Video Order Sensitivity.}
We construct a shuffled-video condition by randomly permuting the order of the frames in the prior video while keeping the frame set, frame count, visual content, and all other inputs unchanged.
\Cref{tab:internvention} shows a significant performance drop across all tasks.
Notably, the decline is most severe on Count and Order—tasks intrinsically dependent on temporal coherence in video trajectories: on Count, SPL, SR, and RSR drop by 13.2\%, 20.0\%, and 21.2\%, respectively; on Order, the corresponding decreases are 16.2\%, 24.0\%, and 24.8\%. 
In contrast, Distance and Orientation—tasks primarily grounded in static geometric relationships—exhibit comparatively modest declines: on Distance, SPL, SR, and RSR decrease by 6.3\%, 9.2\%, and 11.0\%; on Orientation, the drops are 5.3\%, 8.8\%, and 5.6\%. 
We conjecture that while frame shuffling eliminates temporal cues, the preserved visual content and overlapping regions across frames still supports geometric reasoning. 
By contrast, Instance—a task requiring precise entity-grounded resoning—suffers nontrivial degradation (SPL, SR, and RSR fall by 11.8\%, 17.6\%, and 16.4\%), indicating that fine-grained positional reasoning between objects and their immediate surroundings depends on intact spatiotemporal structure, which is disrupted when frame order is randomized.

\subsection{Qualitative Results of MV-DualVLN}

\begin{figure}[h]
    \centering
    \includegraphics[width=0.8\linewidth]{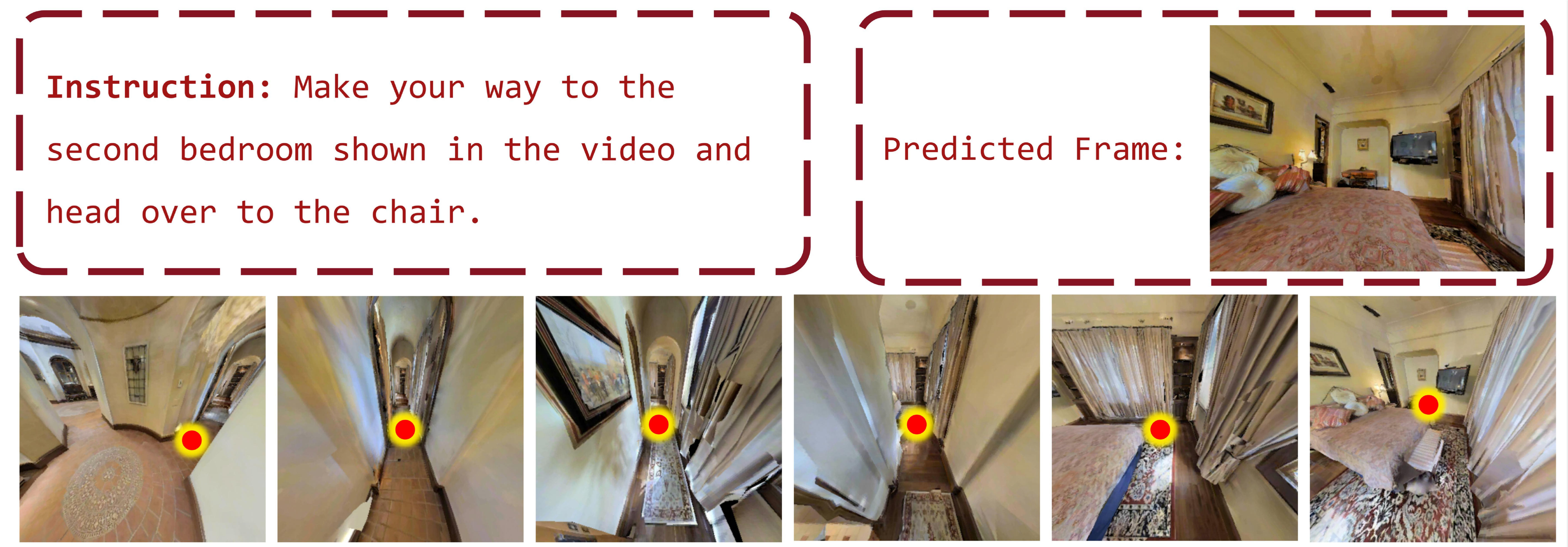}
    \caption{MV-DualVLN visualization example.}
    \label{fig:vsnav1}
\end{figure}

\begin{figure}[h]
    \centering
    \includegraphics[width=0.8\linewidth]{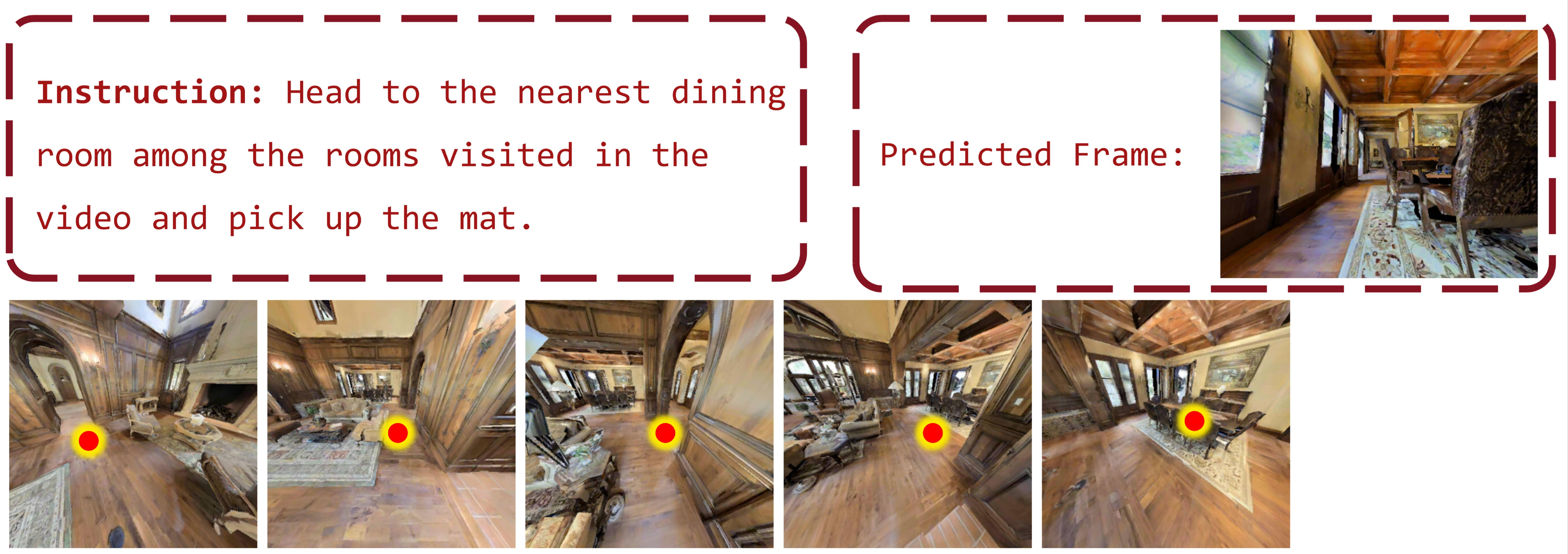}
    \caption{MV-DualVLN visualization example.}
    \label{fig:vsnav2}
\end{figure}

\begin{figure}[h]
    \centering
    \includegraphics[width=0.8\linewidth]{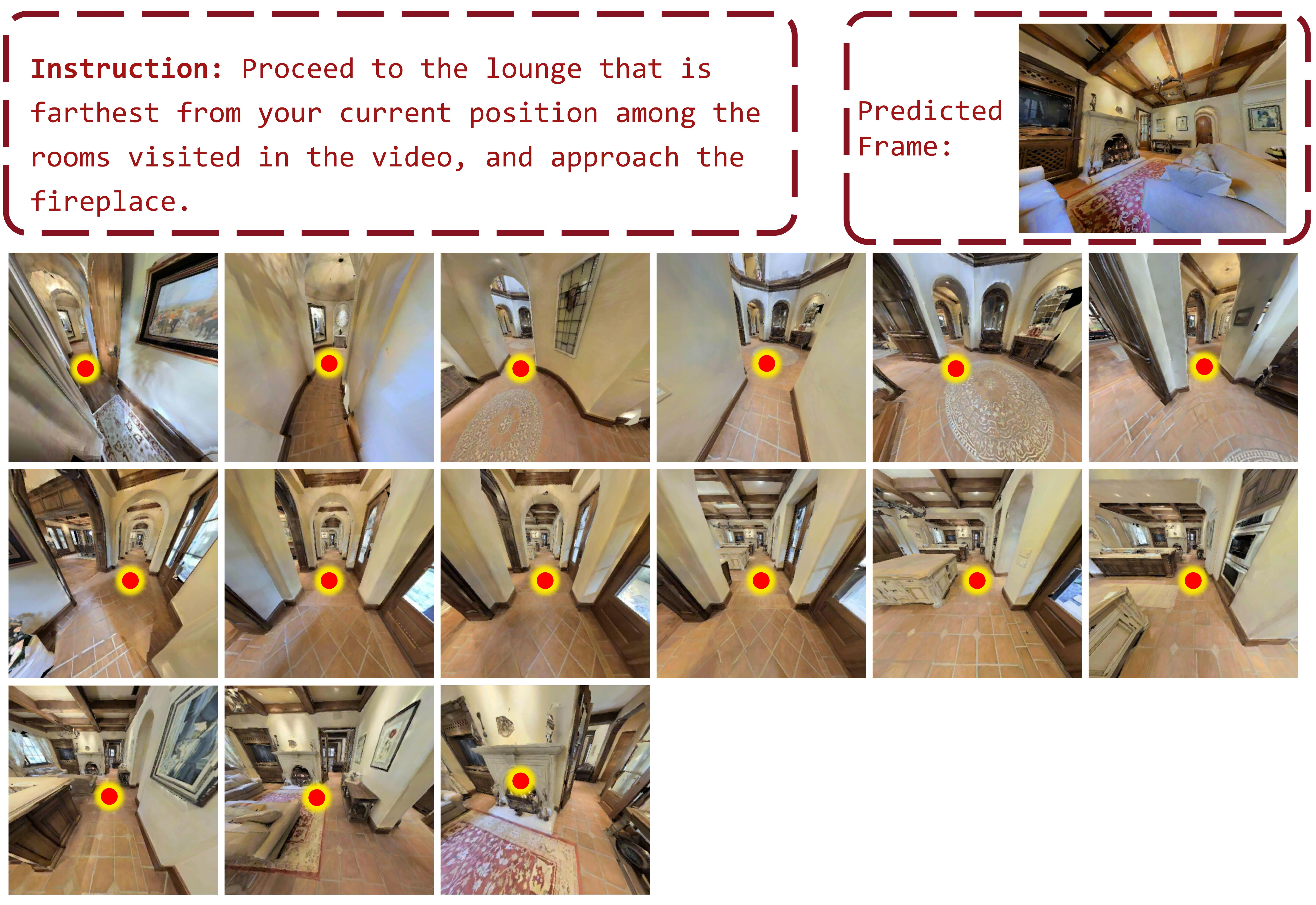}
    \caption{MV-DualVLN visualization example.}
    \label{fig:vsnav3}
\end{figure}

We present three visualization samples of MV-DualVLN results in \Cref{fig:vsnav1,fig:vsnav2,fig:vsnav3}. As shown, during navigation, MV-DualVLN clearly determines the next waypoint via pixel-goal and locates the target object in the final step.


\section{Limitations and Future Work}
% Firstly, several navigation baselines, including MV-DualVLN, use oracle low-level execution. This design factors out control failures and allows the analysis to focus on spatial reasoning and high-level navigation decisions, but it represents an idealized execution setting. Therefore, their results cannot fully reflect the challenges of physical robot deployment.

Firstly, since we focus on evaluating spatial reasoning capabilities grounded in prior visual experience, VCN-Bench is constructed from static Matterport3D scans and assumes that the scene remains unchanged between the prior-video and navigation phases. Real environments may contain moving people, displaced objects, and temporary obstacles. 
Most VCN-Bench episodes focus on Room-to-Object goals and room-level relations that are relatively stable over time; nevertheless, environmental changes may invalidate object-level evidence and alter local traversability. Handling such changes requires agents to detect outdated context and revise their decisions accordingly, which is outside the scope of the current benchmark.
% Under such changes, an agent must not only reason over prior observations but also detect when the prior context has become outdated and revise its decisions accordingly. These capabilities are outside the scope of the current benchmark.

Secondly, our model coverage remains limited, particularly for proprietary MLLMs, due to API costs and heterogeneous visual-input constraints. Some APIs accept substantially fewer frames than the average prior video in VCN-Bench, and aggressive downsampling may remove destination evidence or disrupt the trajectory structure required by the instructions.

Finally, the prior video in VCN-Bench comprehensively covers the navigation initial position and target destination. However, real-world scenarios may provide only partial environmental observations—requiring agents to perform efficient exploration under information constraints.

In future work, we will extend task design to rigorously evaluate models’ reasoning over both known and unknown spatial regions under partial environmental priors. 
At model level, we will investigate diverse visual experience representations—including textual scene descriptions and structured scene graphs—to enhance memory compression and spatial understanding.

\section{Social Impact}
Our work presents VCN-Bench, a foundational benchmark for evaluating closed-loop spatial reasoning over prior visual experience of MLLMs in embodied navigation. It primarily drives progress in multimodal embodied intelligence and has clear positive societal values, while also carrying potential limited negative impacts that require attention and mitigation.

Advances in spatial intelligence evaluated by VCN-Bench directly benefit home service robots, assistive robots for the elderly and people with disabilities, and indoor autonomous devices. By improving MLLMs’ spatial reasoning and navigation capabilities, these robots can better perform daily assistance tasks (e.g., fetching objects, indoor guidance), enhancing quality of life and social inclusivity.

However, since VCN-Bench is built on Matterport3D, which mainly covers Western-style indoor scenes. This bias may lead MLLMs trained/evaluated on this benchmark to perform poorly in non-Western architectural layouts, accessible housing, or minority living environments, creating unfair navigation service for specific groups.

\section{License and Access}
As mentioned in the main paper, our real-world scene data is sourced from Matterport3D~\citep{chang2017matterport3d}. To access and use these datasets, users should follow the original licenses~\url{https://kaldir.vc.in.tum.de/matterport/MP_TOS.pdf}, and ask their official hosts for authorization. Additionally, our annotated data comes from
EmbodiedScan~\citep{wang2024embodiedscan}, access to these datasets requires submitting a request via a Google
Form~\url{https://docs.google.com/forms/d/e/1FAIpQLScUXEDTksGiqHZp31j7Zp7zlCNV7p_08uViwP_Nbzfn3g6hhw/viewform} and following the license attached to the form.

\end{document}